\documentclass[transmag]{IEEEtran}
\usepackage{latexsym}
\def\BibTeX{{\rm B\kern-.05em{\sc i\kern-.025em b}\kern-.08em T\kern-.1667em\lower.7ex\hbox{E}\kern-.125emX}}

\usepackage{cite}
\usepackage{amsmath,amssymb,amsfonts}
\usepackage[noend]{algorithmic}
\usepackage{algorithm}
\usepackage{graphicx}
\usepackage{textcomp}
\usepackage{subcaption}
\usepackage{textcomp}
\usepackage{xcolor}
\usepackage{mathtools}
\usepackage{tabularx}
\usepackage{makecell}
\usepackage{dsfont}
\usepackage[absolute,overlay]{textpos}
\usepackage{comment}
\usepackage{hyperref}
\usepackage{lastpage}
\hypersetup{
    colorlinks=true,
    linkcolor=black,
    filecolor=black,      
    urlcolor=black,
    citecolor=black,
}

\def\etal{\emph{~et~al. }}
\DeclarePairedDelimiter\abs{\lvert}{\rvert}

\begin{document}

%%%%%%%%%%%%%%%%%%%%%%%%%%%%%%%%%%%%%%%%%%%%%%%%%%%%%%%%%%%%%%%%%%%%%%%%%%%%%%%%
%%%%%%%%%%%%%%%%%%%%%%%%%%%%%%%%%%%%%%%%%%%%%%%%%%%%%%%%%%%%%%%%%%%%%%%%%%%%%%%%
\title{Robustness and Adaptability of Reinforcement Learning based Cooperative Autonomous Driving in Mixed-autonomy Traffic}
% Robustifying Cooperative Autonomous Vehicles to Heterogeneity of Human Driver Behaviors
% Robustness and Adaptability of Cooperative Autonomous Vehicles to Heterogeneity of Human Driver Behaviors
% \title{Robustifying Cooperative Autonomous Vehicles to Heterogeneous Driver Behaviors}
% Robustness and adaptability of Reinforcement learning based Cooperative Autonomous Driving in Heterogeneous Mixed-autonomy Traffic 
%\title{\LARGE \bf Safe and Robust Cooperative Driving in Mixed-Autonomy Traffic}
% \title{\LARGE \bf Robust Altruistic Autonomous Vehicles in Mixed Autonomy Traffic }
% \title{\LARGE \bf Effect of Human Behaviors in Altruistic Cooperative Autonomous Vehicles }
% \title{\LARGE \bf Robust Altruistic Autonomous Vehicles in Mixed Autonomy Traffic }
% \title{\LARGE \bf Adaptability of Reinforcement learning based Cooperative Autonomous Driving in Heterogeneous Mixed-autonomy Traffic}
% \title{\LARGE \bf How to Be Nice When Human Drivers Are Not}
%%%%%%%%%%%%%%%%%%%%%%%%%%%%%%%%%%%%%%%%%%%%%%%%%%%%%%%%%%%%%%%%%%%%%%%%%%%%%%%%
%%%%%%%%%%%%%%%%%%%%%%%%%%%%%%%%%%%%%%%%%%%%%%%%%%%%%%%%%%%%%%%%%%%%%%%%%%%%%%%%
\author{Rodolfo Valiente$^{1}$, Behrad Toghi$^{1}$, Ramtin Pedarsani$^{2}$, Yaser P. Fallah$^{1}$% <-this % stops a space
\thanks{*This material is based upon work partially supported by the National Science Foundation under Grant No. CNS-1932037.}% <-this % stops a space
\thanks{$^{1}$ Rodolfo Valiente, Behrad Toghi, and Yaser P. Fallah are with the Department of Electrical and Computer Engineering, University of Central Florida, Orlando, FL, USA. \tt\small {rvalienter90@knights.ucf.edu}}%
\thanks{$^{2}$ Ramtin Pedarsani is with the Department of Electrical and Computer Engineering, University of California, Santa Barbara, CA, USA.}}%
%%%%%%%%%%%%%%%%%%%%%%%%%%%%%%%%%%%%%%%%%%%%%%%%%%%%%%%%%%%%%%%%%%%%%%%%%%%%%%%%
%%%%%%%%%%%%%%%%%%%%%%%%%%%%%%%%%%%%%%%%%%%%%%%%%%%%%%%%%%%%%%%%%%%%%%%%%%%%%%%%
\IEEEtitleabstractindextext{
\begin{abstract}\label{sec:abstract} 
% 1)---Contextualization, problem
% 2)---Gap: Open Questions  and challenges
% 3)---Show the State-of-the-Art Limitations
% 4)---State the solution
% 5)---Results and State the importance of your study
Building autonomous vehicles (AVs) is a complex problem, but enabling them to operate in the real world where they will be surrounded by human-driven vehicles (HVs) is extremely challenging. Prior works have shown the possibilities of creating inter-agent cooperation between a group of AVs that follow a social utility. Such altruistic AVs can form alliances and affect the behavior of HVs to achieve socially desirable outcomes. We identify two major challenges in the co-existence of AVs and HVs. First, social preferences and individual traits of a given human driver, e.g., selflessness and aggressiveness are unknown to an AV, and it is almost impossible to infer them in real-time during a short AV-HV interaction. Second, contrary to AVs that are expected to follow a policy, HVs do not necessarily follow a stationary policy and therefore are extremely hard to predict. To alleviate the above-mentioned challenges, we formulate the mixed-autonomy problem as a multi-agent reinforcement learning (MARL) problem and propose a decentralized framework and reward function for training cooperative AVs. Our approach enables AVs to learn the decision-making of HVs implicitly from experience, optimizes for a social utility while prioritizing safety and allowing adaptability; robustifying altruistic AVs to different human behaviors and constraining them to a safe action space. Finally, we investigate the robustness, safety and sensitivity of AVs to various HVs behavioral traits and present the settings in which the AVs can learn cooperative policies that are adaptable to different situations.
\end{abstract}
%%%%%%%%%%%%%%%%%%%%%%%%%%%%%%%%%%%%%%%%%%%%%%%%%%%%%%%%%%%%%%%%%%%%%%%%%%%%%%%%
%%%%%%%%%%%%%%%%%%%%%%%%%%%%%%%%%%%%%%%%%%%%%%%%%%%%%%%%%%%%%%%%%%%%%%%%%%%%%%%%
}

\maketitle

%%%%%%%%%%%%%%%%%%%%%%%%%%%%%%%%%%%%%%%%%%%%%%%%%%%%%%%%%%%%%%%%%%%%%%%%%%%%%%%%
%%%%%%%%%%%%%%%%%%%%%%%%%%%%%%%%%%%%%%%%%%%%%%%%%%%%%%%%%%%%%%%%%%%%%%%%%%%%%%%%
\section{INTRODUCTION}
\label{sec:introduction}
% 1)---Contextualization: Present the research field and show the importance
% Introducing the need for AVs 
% AVs will impact safety and efficiency, CAVs are important because of communication allowing coordination, CAVs allow is not enough because AVs are not alone, mixed-autonomy is hard because of humans
\IEEEPARstart{T}{he} development of autonomous vehicles (AVs) is on the verge of passing beyond the laboratory and simulation tests and is shifting towards addressing the challenges that limit their practicality in today's society. While there is still need for further technological improvements to enable safe and smooth operation of a single AV, a great deal of research attention is being focused on the emerging challenge of operating multiple AVs and the co-existence of AVs and human-driven vehicles (HVs)~\cite{schwarting2019social, toghi2021social}. A realistic outlook for the adoption of autonomous vehicles on the roads is a mixed-traffic scenario in which human drivers with different driving styles and social preferences share the road with AVs that are perhaps built by different manufacturers and hence follow different policies~\cite{toghi2021cooperativearxiv, sadigh2018planning}. In this work, we seek a solution that can ensure the safety and robustness of AVs in the presence of human drivers with heterogeneous behavioral traits. 

% 2)---What is the problem? Gap: Open Questions, Restrictions and Limitations 
%coordination is necessary , egoism difficult coordination, altruism is important, furthermore need to prioritize safety and robustness
Connected \& autonomous vehicles (CAVs) via vehicle-to-vehicle (V2V) communication allow vehicles to directly communicate with their neighbors, creating an extended perception that enables explicit coordination among vehicles to overcome the limitations of an isolated agent~\cite{shah2020rve,toghi2018multiple, toghi2019analysis, toghi2019spatio, shah2019real, saifuddin2020performance,razzaghpour2021impact}. While planning in a fully AV scenario is relatively easy to achieve, coordination in the presence of HVs is a significantly more challenging task, as the AVs not only need to react to road objects but also need to consider the behaviors of HVs~\cite{aoki2020cooperative, toghi2021cooperativearxiv, sadigh2018planning}. We start by identifying the major challenges in the domain of behavior planning and prediction for AVs in mixed-autonomy traffic. As a preliminary, it is important to distinguish between the individual traits of a human driver, e.g., aggressiveness, conservativeness, risk-tolerance, and their social preferences, e.g., egoism and altruism. Despite the correlation between the two categories, they arise from different natures and also lead to different behaviors in mixed traffic. As an example, an aggressive driver is not necessarily egoistic and selfish, but their aggression might limit their capability to cooperate with other drivers and take part in a socially desirable co-existence with AVs~\cite{sagberg2015review,harris2014prosocial,vallieres2014intentionality}. First, one major challenge is that human drivers are heterogeneous in their individual traits and social preferences, which makes the autonomous vehicle behavior planning extremely difficult, as it is challenging for the AV to predict the type of behavior it is going to face when dealing with a human driver. Additionally, relying on real-time inference of HVs' behaviors is not always viable as the interaction time between vehicles can be short-lived, e.g., two vehicles that meet in an intersection. Second, the driving task involves complex interactions of agents in a partially observable and non-stationary environment, as HVs do not follow a stationary policy and change their policies in real-time according to the other vehicles' behaviors.
% and AVs also changes their policies during the learning process.  

%
\begin{figure*}[t]
  \centering
  \includegraphics[width=.99\textwidth]{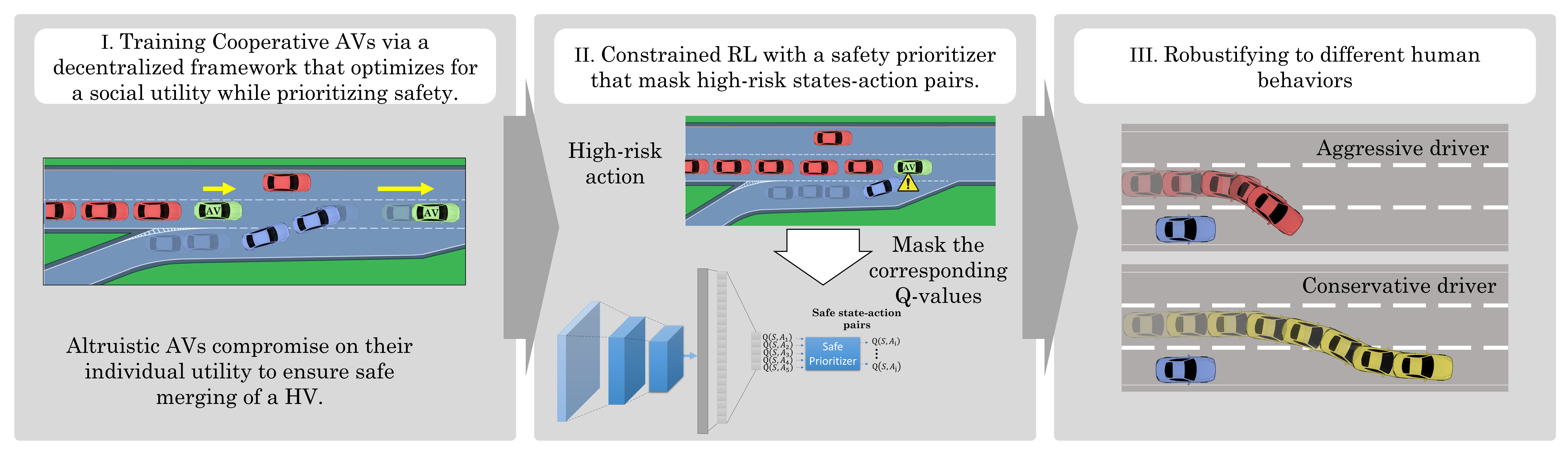}
  \caption{\small{An overview of our methodology for ensuring the safety and robustness of cooperative autonomous vehicles in interaction with human-driven vehicles.}}
  \label{fig:cover}
\end{figure*}
%

% 3)---Show the State-of-the-Art Limitations What are the solutions? SoTA-prior 
% Classical decision-making is impractical in complex scenarios
% 4)---What is unsolved?
% Robustness and safety 
In a pursuit to alleviate the challenges of this co-existence and enable social navigation for AVs, existing works either rely on models of human behavior derived from pre-recorded driving datasets~\cite{sadigh2016planning, wu2017emergent} or defining social utilities that can enforce a cooperative behavior among AVs and HVs~\cite{toghi2021social, toghi2021altruisticarxiv}.
Other works focus on rule-based methods that use heuristics and hand-coded rules to guide the AVs~\cite{rios2016survey} or probabilistic driver modeling~\cite{8690570, 8690965, mahjoub2019v2x}. While this is feasible for simple situations, these methods become impractical in complex scenarios.
The majority of the existing literature relies on simulated environments or human-in-the-loop simulations, which limits the capabilities of modeling the interactions of human drivers with AVs and implementing the heterogeneity of human behaviors. This shortcoming hinders the applicability of the resulting solutions as they are often limited to the human behaviors with which they have interacted during the training. In order to accommodate for this, some of the proposed policies in the literature take an overly-conservative approach when interacting with humans~\cite{li2018safe}. Not only this approach leaves the AVs vulnerable to other aggressive drivers, especially in competitive scenarios such as intersections, but can also cause traffic congestion and potential safety threats~\cite{cosgun2017towards, schwarting2019social}.

% 5)---Solution, How to solve: your intuition and idea?
This study builds on our prior work in~\cite{toghi2021cooperativearxiv} and aim to develop a safe and robust training regimen that enables the AVs to work together and influence the behavior of human drivers to create socially desirable traffic outcomes, regardless of humans' driving individual traits and social preferences. Furthermore, we emphasize the importance of safety in social settings and constrain the AVs' policies to remove high-risk actions that can cause safety threats. Our work in this paper is built on the following key insights. 
First, we rely on learning from experience in a decentralized reinforcement learning framework that optimizes for a social utility, and exposes the learning agents to a wide spectrum of driver behaviors. By doing so, the resulting agents become robust to the behavior of human drivers and are able to handle cooperative-competitive behaviors regardless of HV's level of aggression and social preference.
Second, a safety prioritizer is proposed to avoid high-risk actions that can undermine driving safety. The overview of our methodology is presented in Figure~\ref{fig:cover}.

\begin{figure}[b]
  \centering
  \includegraphics[width=.48\textwidth]{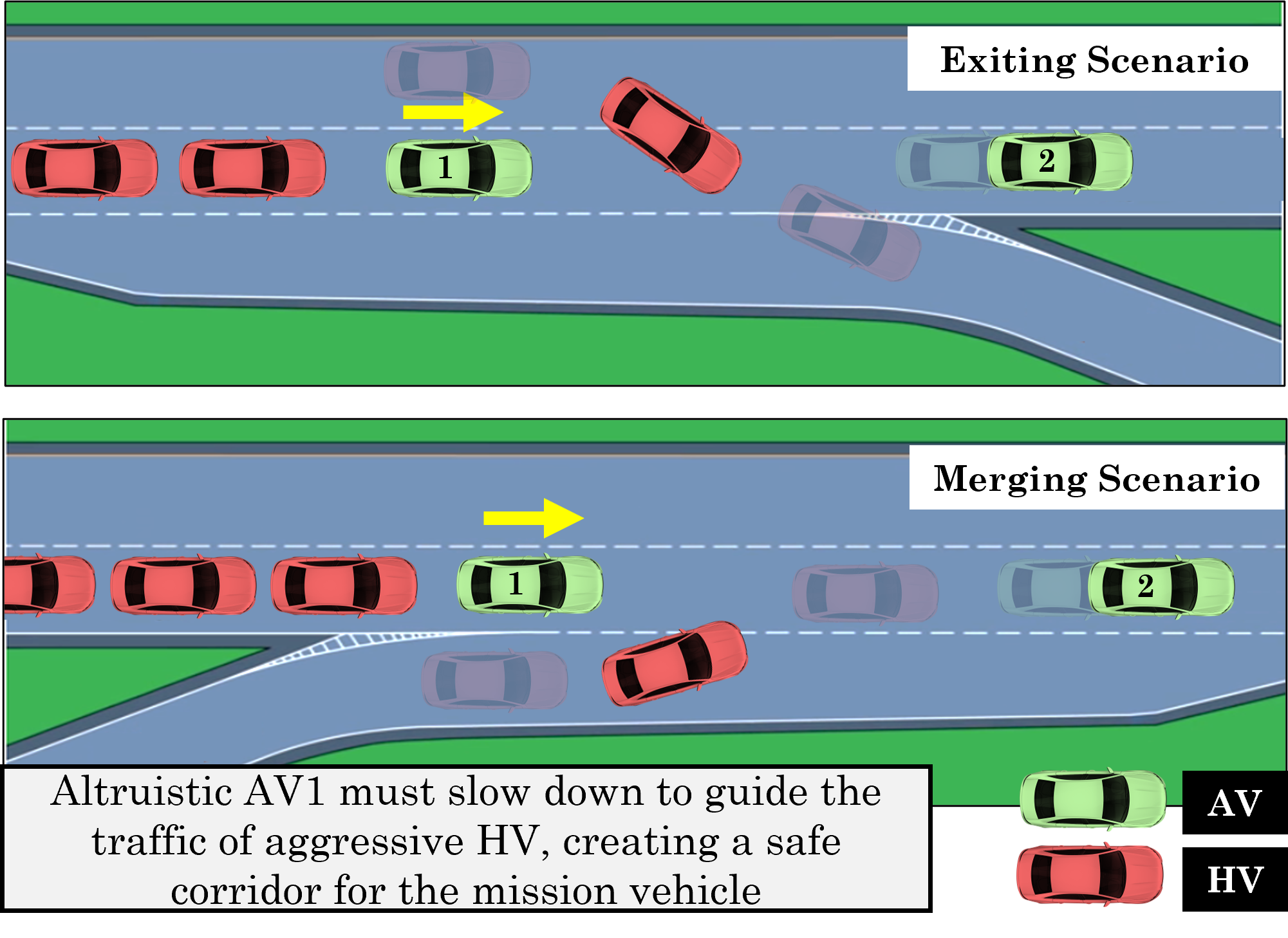}
  \caption{\small{Highway merging and exiting scenarios in mixed traffic where the road is shared by AVs (green) and aggressive HVs (red). Altruistic AVs must learn to coordinate to allow for a safe and efficient merging/exiting while also being robust to different scenarios and behaviors and ensuring safety in decision-making.}}
\label{fig:mainfigure}
\end{figure}

% 7)---Briefly give your principal results % Results and State the importance of your study
%We exploit domain adaptation and transfer learning to foster generalization while efficiently learning harder tasks from trained models and therefore accelerate the learning. 
%Comprehensive experimental results show the proposed MARL framework outperforms state-of-the-art works, allowing AVs to adapt to diverse HVs behaviors and traffic scenarios and demonstrating a significant improvement in both safety and traffic efficiency as a result of this prosocial safe behavior.
Ultimately, the focus of this paper is on exploring the safe and robust decision-making problem in a mixed-autonomy MARL environment, in inherently competitive driving scenarios, such as the ones illustrated in Figure~\ref{fig:mainfigure}, where cooperation is required for safety and efficiency. The purpose is not to fully solve the autonomous driving problem, but instead, use it as an example to investigate the effectiveness of societal concepts from psychology literature within the MARL domain. Further work is required to use these ideas on real roads. Nevertheless, we are encouraged to see altruistic AVs that are safe, robust, and can learn to influence humans in a desired way without the limitations of current solutions
\cite{valiente2020dynamic,valiente2019controlling,valiente2020connected,toghi2021cooperativearxiv, toghi2020maneuver, toghi2021towards}.
% 6)---Key contributions!!

Our main contributions are:
\begin{itemize}
  \item We begin by formulating the mixed-autonomy traffic as a stochastic game and introduce a general decentralized framework for training cooperative AV, that optimizes for a social utility while prioritizing safety and allowing adaptability. 
  \item A novel training regimen is introduced that robustifies the AVs' capability in creating socially desirable outcomes with regards to human drivers' behavior.
  \item We proposed a safety prioritizer that constrained the policy of cooperative AVs to ensure the safety of their behavior via masking the Q-states that lead to high-risk outcomes.
\end{itemize}

%---------------------------------------------------------------------
% Main fig, Idea for Fig. 1: Consider Altruism, behavior/robustness sensitivity, and adaptation. AVs learn from moderate HVs behavior and adapt to conservative HVs. On a conservative behavior, the decision is generally accelerating, in an aggressive environment, the decision is guiding the aggressive HVs to allow a safe merging/exit. Our approach results in fewer collisions and improved navigation in complex scenarios. In addition, adaptation/generalization from merging/to exit and vice versa is necessary toward a more general solution.It's not sufficient to be altruistic, AVs should also be robust and prioritize safety.
%---------------------------------------------------------------------

%%%%%%%%%%%%%%%%%%%%%%%%%%%%%%%%%%%%%%%%%%%%%%%%%%%%%%%%%%%%%%%%%%%%%%%%%%%%%%%%
%%%%%%%%%%%%%%%%%%%%%%%%%%%%%%%%%%%%%%%%%%%%%%%%%%%%%%%%%%%%%%%%%%%%%%%%%%%%%%%%
\section{Literature Review}
\label{sec:relatedworks}
% 1)---Introduce section ?? What we will cover?
MARL has received a lot of attention from the research community in recent years. MARL algorithms that assume separately trained agents perform poorly due to the intrinsic non-stationarity of the environment~\cite{hernandez2017survey}. Some efforts to solve this issue assume that all agents observe the global state~\cite{chu2019multi} or that they can share their states with their neighbors~\cite{arel2010reinforcement}. These assumptions address the non-stationarity challenge; however, they are unfeasible on real roads\cite{oroojlooyjadid2019review}. Authors in~\cite{foerster2018counterfactual,gupta2017cooperative,lowe2017multi} take steps to address this challenge and~\cite{foerster2018counterfactual} propose a centralized critic that reduces the influence of non-stationarity during the learning process. Xie~\etal consider a MARL approach that learns latent representations of the agent's policies, modeling agent strategies that depend on long interactions, alleviating the non-stationary effect, and enabling better performance and co-adaptation~\cite{xie2020learning}. ~\cite{shih2021critical} further investigates the impact of interactions on agent's modeling. Authors in~\cite{vinitsky2021learning} present a RL agent that learns to acquire social norms from public sanctions using a decentralized framework.

\subsection{Driver Behavior and Social Navigation}
Social navigation in mixed autonomy has shown the potential of collaboration among AVs and HVs~\cite{pokle2019deep}.
Current works in social navigation tackle the MARL cooperation by assuming the nature of agent interactions~\cite{lauer2000algorithm,omidshafiei2017deep} or by directly modeling or classifying human driver behaviors~\cite{brown2020taxonomy,ivanovicgenerative,mahjoub2019representing}. Different methods to predict or classify driver behaviors are based on driver attributes~\cite{beck2014distress}, graph theory~\cite{chandra2020cmetric}, game theory~\cite{schwarting2019social} and data mining~\cite{constantinescu2010driving}. Toghi~\etal release a maneuver-based dataset and propose a model that can be used to classify driving maneuvers~\cite{toghi2020maneuver}. Authors in~\cite{ivanovicgenerative} present an approach to modeling and predicting human behavior in situations with several humans interacting in highly multi-modal scenarios that could allow AVs to predict human reactions.
\cite{brown2020taxonomy} can be referred for a comprehensive study on modeling and prediction in multi-agent traffic scenarios.

In~\cite{kuderer2015learning} and~\cite{sadigh2018planning} the driving patterns of humans are learned from demonstration through inverse RL. An approach based on imitation learning is proposed by~\cite{sadigh2016planning} to learn a reward function for human drivers and demonstrates how AVs can manipulate human behaviors. In~\cite{hadfield2016cooperative} a centralized game-theory model for cooperative inverse reinforcement learning is proposed. Several works take a more abstract and traffic-level perspective~\cite{wu2018stabilizing, vinitsky2020optimizing,lazar2021learning}. Differently, we rely on implicitly learning from experience altruistic behaviors that facilitate AVs' coordination without the need for a human model or counting on their collaboration.

\subsection{Safe and Robust Decision-Making}
% to handle the diversity of real traffic situations
In addition to the socially advantageous behavior of altruistic AVs, it is important to consider robustness and safety. 
Safety is essential for AVs and is particularly important for AVs trained using RL. We need to prioritize safety; as coordination is often coupled with risk, in many situations there exists a safe action that produces lower rewards, and a risky action that produces higher rewards if agents coordinate~\cite{biyik2018altruistic,wang2021emergent}, however, the risky action increases the probability of crashes when synchronization fails. In particular, AVs using trained RL algorithms may not always behave safely as the trained models may choose unsafe actions~\cite{li2018safe}.  In that direction, several works take a pure reward shaping approach to avoid collisions. While this is a common practice in RL, safety is not implicitly prioritized and AVs using those RL algorithms may not behave safely in some scenarios, as the agents could choose dangerous actions due to function approximation.

To address this challenge, the idea of safe RL is proposed in~\cite{li2018safe} to improve safety in unseen driving environments in which the RL algorithm behaves unsafely. ~\cite{wang2019lane} presents a rule-based decision-making framework that examines the trajectories given by the controller and replaces the actions causing collisions. Nageshrao~\etal~\cite{nageshrao2019autonomous} includes a short-horizon safety supervisor to substitute risky actions with safer ones. Nevertheless, these studies consider oversimplified and non-realistic environments. 
The work in~\cite{hoel2019combining,mohammadhasani2021reinforcement} utilizes a Q-masking approach to prevent collisions, removing the actions that could result in a crash. Chen~\etal present a novel priority-based safety supervisor to significantly reduce collisions~\cite{chen2021deep}. 

In this work, we leverage these approaches to improve the safety of the altruistic agents while also training the cooperative agents to be robust to different driver behaviors and scenarios using a decentralized reward function, local actions, and assuming partial observability.
We consider a general setup where AVs and HVs with different behaviors coexist as in Figure~\ref{fig:mainfigure}. Particularly, the figure represents two common traffic situations where vehicles are either required to efficiently merge to a lane or exit the highway without colliding with other vehicles. In an ideal cooperative setting, the vehicles should proactively decelerate or accelerate to make adequate space for the vehicles to safely merge/exit and avoid deadlock situations, while also being robust to different scenarios and behaviors and ensuring safety in decision-making.
 
\section{Problem Formulation}
\label{sec:problem_formulation}
% \section{Problem Statement}
% \label{sec:problem_statement}
% 1)---In this section we formally define our problem of interest

%---------------------------------------------------------------------
% TODO Long version reduce
% \subsection{Partially Observable Stochastic Games (POSG)}
% As discussed, cooperative MARL overcome the restrictions of a single agent. 
% We can formalize the MARL problem as a centralized or decentralized problem. 
The MARL problem can be formulated as a centralized or decentralized problem. A centralized controller that assigns a central joint reward and joint action is straightforward. Nevertheless, such assumptions are impractical in the real world. Therefore, this study focuses on a decentralized controller where agents have partial observability and consequently, the problem is formulated as a partially observable stochastic game (POSG) defined by $\langle \mathcal{I}, \mathcal{S}, \{ \mathcal{A}_i \}_{i\in\{1,...,N\}}, \{ \mathcal{O}_i \}_{i\in\{1,...,N\}}, P, \{ R_i \}, \gamma \rangle$ where
\begin{itemize}
\item $\mathcal{I}$: a finite set of agents $N \geq 2$. %($N=1$ degenerates to a single-agent MDP).
\item $\mathcal{S}$ : a set of possible states that contains all configurations that $N$ AVs can take (probably infinite).
\item $\mathcal{A}_i$: a set of possible actions for agent $i$.
\item $\mathcal{O}_i$: a set of observations for agent $i$.
% \item $P$: a state transition probability function. %$P_{ss'}^a = P(S=s'|S=s,A=a)$.
\item $P$: a state transition probability function from state $s \in \mathcal{S}$ to state $ s' \in \mathcal{S}$, $P(S=s'|S=s,A=a)$.
% \item $P(S=s'|S=s,A=a)$: a state transition probability function %from state $s \in \mathcal{S}$ to state $ s' \in \mathcal{S}$.
%\item $P$: for each time step $t \in \mathbb{N}$, the transition probability from state $s \in \mathcal{S}$ to state $ s' \in \mathcal{S}$.
\item $R_i$: a reward function for the $i^{th}$ agent, $R_i(s,a)$. % that returns a scalar value to the $i-th$ agent for a transition from $(s,a)$ to $s'$. The rewards is bounded by $R_{\text{max}}$. 
\item $\gamma$: a discount factor, $\gamma \in [0, 1]$. 
\end{itemize}

It should be noted that the agents have no access to the exact environmental state but only a local observation which is correlated with the state, increasing the difficulty of solving the POSG. The POSG can be described as follows: at every time step $t$, $s_t$ is the state of the environment, each agent senses the environment and obtains a local observation $o_i: \mathcal{S} \rightarrow \mathcal{O}_i$, based on $o_i$ 
and its stochastic policy $\pi_i$, it selects an action from the action space $a_i \in \mathcal{A}_i$. As a result, the agent moves to the next state $s'$ and obtains a decentralized reward $r_i: \mathcal{S} \times \mathcal{A}_i \rightarrow \mathbb{R}$. The goal of each agent $i$ is to optimally solve the POSG by deriving a probability distribution over actions in $\mathcal{A}$ at a given state, that maximizes its cumulative discounted reward over an infinite time horizon, and find the corresponding optimal policy $\pi^*: \mathcal{S} \rightarrow \mathcal{A}$.

An optimal policy maximizes the action-value function, i.e., $\pi^*(s) = \arg\max_a Q^* (s,a)$ where $Q^\pi(s,a) \coloneqq \mathbb{E}_{\pi} [\sum_{k=0}^\infty \gamma^k R_k(s,a) |s_0=s, a_0=a]$. The optimal action-value function is determined by solving the Bellman equation $Q^*(s,a) = \mathbb{E} \left[ R(s,a) + \gamma \max_{a'}  Q^*(s',a') |s_0=s, a_0=a \right]$.
%$Q^*(s,a) = \mathbb{E}_{s'\sim P(.|s,a)} [ r(s,a) + \max_a' \gamma Q^*(s',a') ]$.
% \vspace{-8pt}
% \begin{equation}
% \label{equ:bellmanequ}
% Q^*(s,a) = \mathbb{E} \left[ R(s,a) + \gamma \max_{a'}  Q^*(s',a') |s_0=s, a_0=a \right]
% % Q^*(s,a) = \mathbb{E}_{s'\sim P(.|s,a)} \left[ R(s,a) + \gamma \max_{a'}  Q^*(s',a') |s_0=s, a_0=a \right]
% \end{equation}

%---------------------------------------------------------------------

%---------------------------------------------------------------------
% \smallskip

% \noindent \textbf{Double Deep Q-Network (DDQN). }
\subsection{Double Deep Q-Network (DDQN)}
We use Double Deep Q-Network (DDQN) as the function approximator to estimate the action-value function, i.e., $\Tilde{Q}(.;\textbf{w}) \cong Q(.)$ (with weights $\textbf{w}$)~\cite{van2016deep}. DDQN improves Deep Q-Network (DQN) by decomposing the max operation in the target into action selection and action evaluation, mitigating the over-estimation problem. The idea is to periodically sample data from a buffer and compute an estimate of the Bellman error or loss function, written as
% \vspace{-5pt}
\begin{equation}
\label{equ:loss2}
\mathcal{L}(\textbf{w}) = \mathbb{E}_{s,a,r,s' \sim \mathcal{RM}}[( Target - \Tilde{Q}(s,a;\textbf{w}))^2]
\end{equation}
\vspace{-5pt}
\begin{equation}
\label{equ:DDQNtarget}
Target = R(s,a) + \gamma \Tilde{Q}(s',\underset{a'}{\arg\max} \Tilde{Q}(s',a';\textbf{w});\hat{\textbf{w}}))
\end{equation}

% %
The DDQN algorithm then applies mini-batch gradient descent as $\textbf{w}_{k+1} = \textbf{w}_k - \alpha \hat{\nabla}_\textbf{w} \mathcal{L}(\textbf{w})$, on the loss $\mathcal{L}$ to learn the approximation of the value function ($\Tilde{Q}(.)$). The $\hat{\nabla}_\textbf{w}$ operator denotes an estimate of the gradient at $\textbf{w}_k$,  $\textbf{w}$ are the weights of the online network and $\hat{\textbf{w}}$ are the weights of the target network which are updated at a lower frequency ($Target_{update}$).

%---------------------------------------------------------------------

%---------------------------------------------------------------------
% \smallskip

% \noindent \textbf{Driving Scenarios.}
\subsection{Driving Scenarios}
Our goal is to study driving scenarios where the absence of coordination by the AVs compromises safety and efficiency. Additionally, we aim to investigate adaptability among competitive scenarios and driver behaviors. For this purpose, we define a set of scenarios $\mathcal{F}$ and choose highway merging and exiting ramp as the base scenarios where a mission vehicle (merging/exiting) attempts to complete its task in a mixed-traffic environment as in Figure~\ref{fig:mainfigure}. 

We design the merging and exit scenarios such that cooperation is required for safety. AVs must coordinate and neither of them alone can realize a safe and smooth traffic flow, i.e., only one AV will not make the merging/exiting possible without the coordination of the other AVs. The altruistic AVs must learn to account for the interests of all the vehicles, coordinate, make sacrifices and guide the behavior of humans to allow for a safe merging/exiting while also adapting to different traffic situations safely. For instance, in Figure~\ref{fig:mainfigure}, the AV1 must learn to sacrifice and slow down (compromising its own utility) to guide the traffic of aggressive HVs, creating a safe corridor for the merging/exiting vehicle; while the other AVs have to accelerate to make space for the mission vehicle. The merging and exiting scenarios are defined as $f_m , f_e \in \mathcal{F} $ respectively. We select such situations because of their intrinsic closeness and competitive characteristics, since the merging/exiting vehicle's local utility conflicts with that of the highway vehicles. 
%Furthermore, we design our experiment so that necessarily all AVs must work together and neither of them alone can realize a safe and smooth traffic flow, i.e., only one AV will not make the merging/exit possible without the coordination of the other AVs. To investigate adaptability, we tested the leaned policies in the two different scenarios and in the presence of HVs with various level of aggressiveness.
%---------------------------------------------------------------------

%---------------------------------------------------------------------
% \smallskip

% \noindent \textbf{Social Value Orientation and Altruistic AVs.}
\subsection{Social Value Orientation and Altruistic AVs}
% SVO indicates a person’s preference between themselves and another person. 
Social Value Orientation (SVO) characterizes the individual's preference to account for the interests of others vs. their own interest~\cite{schwarting2019social}.
% , in our setting, it can be defined as a degree to which an agent acts egoistic or altruistic in the presence of others. 
The behavior of a human or an AV can fluctuate from egoistic to absolutely altruistic based on the importance given to the utility of others. The SVO of humans is uncertain, therefore we depend on AVs instead to guide the traffic towards more socially advantageous goals. 
Formally, the SVO angle $\phi$ of an AV, determines how the AV balances its own benefit against that of others. In terms of rewards, we can define the total reward $R_i$ of an AV as: $R_i(s,a) = \cos \phi_i r^{ego}_i + \sin \phi_i r^{social}_i$, in which $r^{ego}_i$ is the AV's specific reward (egoistic) and $r^{social}_i$ is the overall reward of other vehicles (social) respect to the $i^{th}$ AV~\cite{toghi2021cooperativearxiv, toghi2021social}. The SVO angle can be changed from $\phi=0$ (purely egoistic) to $\phi=\pi/2$ (purely altruistic). Nevertheless, none of the two extremes is optimum, and a point in between yields the most socially advantageous result, defined as the optimal SVO angle~$\phi^*$.
% For a comprehensive description of SVO refer to~\cite{schwarting2019social, toghi2021social}.
SVO helps explain the behaviors that allow the mission vehicle to merge or exit in Figure~\ref{fig:mainfigure}. 
Without SVO, the mission vehicle in Figure~\ref{fig:mainfigure} could cause traffic congestion or an unsafe situation. AVs need to consider SVO, since HVs can not communicate that directly, and we should not expect HVs to cooperate. 

%---------------------------------------------------------------------

%---------------------------------------------------------------------

% \noindent \textbf{Driving Behaviors.} 
\subsection{Driving Behaviors}
%While the agents learn from experience and optimize for a social utility, the environment also needs to include heterogeneous driver behaviors.
%, so the adaptability of the agents can be tested in the presence of heterogeneous HVs.
The challenge of simulating diverse behaviors can be framed as the problem of obtaining the suitable range of parameters that can generate the heterogeneous behaviors within the simulator. 
Many studies from social traffic psychology establish that driving behavior falls between aggressive and conservative. Nonetheless, the precise definitions differ between studies~\cite{sagberg2015review}. %Authors in~\cite{ sagberg2015review} summarized these studies. 
In general, the term “aggressive driving” covers a range of unsafe driving behaviors like overspeeding or running red lights.
However, the causes of aggressive driving come in various forms and are not always obvious. Some are due to undesirable roadway situations, while others are individual traits or states of mind. 
% Human behaviors are complex and unpredictable. 
Furthermore, there is a  relationship between aggressiveness and egoism, as egoistic drivers usually do not yield and also tend to engage in speeding, risk-taking, and similar aggressive behaviors. While there is a correlation between these terms~\cite{harris2014prosocial,sagberg2015review,vallieres2014intentionality}, for the purpose of this paper, we separate egoism from aggressiveness by characterizing social preferences and individual traits. 

% Properly, for our problem,
% Properly, we consider egoism as a selfish driver that accounts for their own utility independently of their aggressiveness and we consider aggressiveness as a more general term about the drivers’ behavior. For instance, we could imagine selfish people who drive safely and behave conservative because they want to protect themselves.
We differentiate between social preferences and individual traits of drivers, as they lead to different behaviors. First, we characterize egoism and altruism as social preferences, and identify an egoistic driver as a selfish driver who accounts for his own utility independently from his aggressiveness. Second, we characterize aggressiveness and conservativeness as individual traits, and identify an aggressive driver as a driver whose actions cause aggressive behaviors. Social preferences such as egoism are characterized by their social goals and intentions, whereas individual traits such as aggressiveness are characterized by the consequences of their actions. In that sense, an egoistic driver is a self-centered driver characterized by a lack of social motivation, a driver that feels like he owns the road and does not consider the other drivers. Egoist drivers often engage in aggressive behaviors and while ego defensiveness is not the only cause of aggressiveness, it is still a main contributing factor of anger and aggressive driving~\cite{harris2014prosocial,vallieres2014intentionality}. Despite the correlation between the two categories, they arise from different natures and lead to different behaviors.
%In that sense an egoistic agent is a self-centered driver, a driver that feels like they own the road and do not consider the other drivers. Egoism is characterized by a lack of social motivations, egoistic agents are selfish drivers.
For instance, a driver could be egoistic and conservative. We could imagine a driver who drives cautiously in order to protect himself (selfish motivation/preference) and, as a consequence, behaves conservatively (outcome of his actions).

Formally, in our simulation, social preferences (egoism or altruism) are characterized by the AV's SVO angular preference $\phi$; and individual traits (aggressiveness, conservativeness, etc) by the HV driver model parameters ($\mathcal{P}$) as described in section~\ref{sec:humandrivermodel}. Based on the values of these parameters, a vehicle will exhibit aggressive or conservative behaviors. In our experiments, we assume the SVO of HVs to be unknown as they can not communicate that directly.
%the vehicle's behaviors. 
Finally, we define a set of behaviors $\mathcal{B}$, i.e, aggressive, moderate and conservative, $b_a,b_m,b_c \in \mathcal{B}$ based on the parameters ($\mathcal{P}$) obtained in section~\ref{sec:humandrivermodel}.
%---------------------------------------------------------------------

% \noindent \textbf{Problem Formulation.}
\subsection{Problem Formulation}
%We will also show that prosocial policies learn in a merging scenario are transferable to an exit situation in the presence of HVs.
% 2)---Problem Formulation
%------------------------------------------
% Problem (what?): driving in mixed-autonomy is hard, currently AV approaches to solve that problem are inefficient, no robust and unsafe. problem is how to drive in a mixed-autonomy environment, in a robust/adaptable and safe manner. 
% Goal (solve the problem): Our goal is to train AVs that learn how to drive in a mixed-autonomy scenario in a robust, efficient and safe manner while benefiting all the vehicles on the road.
% Solution(how?), using MARL with a general decentralized reward function that optimize for a social utility, by inducing altruism in our agents (SVO, social utility) (provides efficiency); our general reward accounts for any possible vehicle’s mission/pre-desired task, allowing to be applied to different scenarios (provides robustness); and ensure safety by adding a safety prioritizer (provides safety). The final solution is efficient, robust and safe
%------------------------------------------
We formulate the problem as the POSG defined; where the road is shared by a set of HVs $h_k \in \mathcal{H}$, with an undetermined SVO $\phi_k$ and heterogeneous behaviors $b_k \in \mathcal{B}$; a set AVs $i_i \in \mathcal{I}$, that are connected together using V2V communication, controlled by a decentralized policy and sharing the same SVO, and a \emph{mission vehicle}, $M \in \mathcal{I} \cup \mathcal{H}$ that is aiming to accomplish its mission (highway merging/exiting) and can be AV or HV. We focus on the multi-agent maneuver-level decision-making problem for AVs in mixed-autonomy environments, and study the following problems: how AVs can learn in a mixed-autonomy environment cooperative optimal policies $\pi^*(s)$ that are robust to different scenarios $f \in \mathcal{F}$ and behaviors $b \in \mathcal{B}$ while ensuring safety on the decision-making, and how sensitive is the performance of the altruistic AVs to the HVs behaviors. 

As AVs are connected, we assume that they receive an accurate local observation of the environment $\Tilde{\textbf{o}}_{i} \in \widetilde{\mathcal{O}}_i$, sensing all the vehicles within their perception range, i.e, a subgroup of HVs $\widetilde{\mathcal{H}} \subset \mathcal{H}$ and a subgroup of AVs $\widetilde{\mathcal{I}} \subset \mathcal{I}$. Nevertheless, AVs are unable to share their actions or rewards, and they take individual actions from a set of high-level actions $a_i \in \mathcal{A}_i (|\mathcal{A}_i|=5)$. The goal of this work is to train AVs that learn how to drive in a mixed-autonomy scenario in a robust, efficient and safe manner while benefiting all the vehicles on the road. 
%Our goal is to find a safe decentralized control structure that will lead to AVs behaving in an altruistic and robust manner. 
% ??? optimization problem as the eventual advantageous social outcome with adaptability
% We focus on the multi-agent maneuver-level decision-making problem for AVs in mixed-autonomy environments and study the following problems, how AVs can learn cooperative policies that are robust to different scenarios and behaviors and ensure safety on the decision-making, and how sensitivity are our prosocial AVs to different HVs behaviors. 
% A prosocial AVs that account for the interest of other vehicles and optimize for a social utility can learn to coordinate and affect the behavior of HVs leading to socially desirable actions. Furthermore, adaptability is also advantageous as we desire that policies learned in one scenario in the presence of HVS with different behaviors could be relevant to other scenarios. Finally, safety is the most basic requirement in autonomous navigation, and a RL algorithm that do not prioritize safety can lead to undesirable accidents.

%% ? Do we have now a clear understanding of what is exactly the problem, and the formal definition ??****

%%%%%%%%%%%%%%%%%%%%%%%%%%%%%%%%%%%%%%%%%%%%%%%%%%%%%%%%%%%%%%%%%%%%%%%%%%%%%%%%
%%%%%%%%%%%%%%%%%%%%%%%%%%%%%%%%%%%%%%%%%%%%%%%%%%%%%%%%%%%%%%%%%%%%%%%%%%%%%%%%

% \section{Proposed Solution}
\section{Safe and Robust Altruistic Driving}
\label{sec:solution}
% 1)---In this section we present the solution
% We propose an approach in which AVs learn from experience to perform a task, optimizing for a social utility that accounts for the interests of all the vehicles, while being able to adapt to other traffic situations safely.
To drive in a mixed-autonomy environment in a robust and safe manner, we propose a MARL approach with a general decentralized reward function that optimizes for a social utility by inducing altruism in the agents; the general reward accounts for any anticipated vehicle’s mission, allowing it to be applied to different scenarios and tasks; and ensuring safety by adding a safety prioritizer. We train altruistic AVs that learn from experience to perform a task, account for the interests of all the vehicles, while being able to adapt to other traffic situations safely.
% In our approach our AVs learn from experience to perform a task in an efficient, robust, and safe manner
% What we define as "driving" is the outcome of decades of human learning from experience. Consequently, we take the same approach and train AVs that learn from experience and define the optimization problem as the eventual desirable social outcome with adaptability, expecting AVs to learn how to drive safely during the process.Figure~\ref{fig:mainfigure} helps us to create intuition on these points, by introducing driving scenarios in which altruistic AVs lead to socially advantageous results while adapting to different traffic circumstances.
% We develop a safety prioritizer with a decentralized RL algorithm that uses the optimal AVs' SVO value to promotes safe prosocial behavior in their decision-making process.
% We demonstrate that such social safe behavior can be learn through carefully designing of the appropriate action and observation space, a decentralize general reward function, a suitable architecture and safety prioritizer.
We carefully design a decentralized general reward function, a suitable architecture, and a safety prioritizer to promote the desired safe altruistic behavior in AVs' decision-making process. The overview of our approach as presented in Figure~\ref{fig:cover} and Figure~\ref{fig:mainfigure} helps us to create intuition on these points, by introducing driving scenarios in which altruistic AVs lead to socially advantageous results while adapting to different traffic circumstances. 

% \subsection{Action and Observation Spaces}
% \label{sec:spaces}
\textbf{Action Space.}
% \noindent \textbf{Action Space.}
We define a high-level action space $\mathcal{A}$ of discrete meta-actions for decision-making. In particular, we select a set of five high-level actions as $a_i \in \mathcal{A}_i = [\texttt{Change to Right Lane}$, $\texttt{Change to Left Lane}$, $\texttt{Accelerate}$, $\texttt{Decelerate}$, $\texttt{Idle}]^T$. These meta-actions are then converted into trajectories and low-level control signals, which ultimately control the vehicle's movement.

\textbf{Observation Space.}
% \noindent \textbf{Observation Space.}
We use a \textit{multi-channel VelocityMap} observation ($o_i$) that embeds the relative speed of the vehicle with respect to the ego vehicle in pixel values, as in~\cite{toghi2021social}. We represent the information in multiple semantic channels that embed: 1) the AVs, 2) the HVs , 3) the mission vehicle, 4) an attention map to highlight the position of the ego vehicle, and 5) the road layout. To map into pixels the relative speed of the vehicles, we use a clipped logarithmic function which improves the dynamic range and shows better results than a straightforward linear mapping. As temporal information is necessary for safe decision-making, we use a history of VelocityMaps successive observations to create the input state to the Q-network $\psi_i$. 

%%%%%%%%%%%%%%%%%%%%%%%%%%%%%%%%%%%%%%%%%%%%%%%%%%%%%%%%%%%%%%%%%%%%%%%
\subsection{Decentralized General Reward}
\label{sec:reward}
We train the AVs from scratch using local observations and a decentralized reward structure and expect them to learn the driving task in different scenarios while accounting for individual diver's missions. Consequently, we design a well-engineered general reward function that accounts for the social utility, traffic metrics and desired missions.
% Our reward function can work for different scenarios, driving cases and missions without been situation specific. 
The agent's $I_i \in \mathcal{I}$ local reward is defined as
% \vspace{-5pt}
\begin{equation} \label{equ:decentralizedreward}
\begin{aligned}
R_i(s, a) ={} & R^{\mathrm{ego}}+R^{\mathrm{social}}
\\R^{\mathrm{ego}} = {} & \cos \phi_i r_i(s, a)  \\
R^{\mathrm{social}} = &  \sin \phi_i \Big[ \sum_j r^{\mathrm{AV}}_{i, j} (s, a)+ \sum_j r_{i,j}^M (s, a) \\
& + \sum_k r^{\mathrm{HV}}_{i, k} (s, a) + \sum_k r_{i,k}^M (s, a) \Big]\\
\end{aligned}
\end{equation}

in which $R^{\mathrm{ego}}$, $R^{\mathrm{social}}$ represents the egoistic and social reward, $i \in \mathcal{I} $, $j \in (\widetilde{\mathcal{I}} \setminus \{I_i\})$, $k \in \widetilde{\mathcal{H}}$. The term $r_i$ represents the ego vehicle's reward obtained from traffic metrics and the angle $\phi$ allows to adjust the level of egoism or altruism.
We decouple the social component in cooperation (the altruistic behavior among AVs, i.e, AV's altruism toward others AVs) and sympathy (AV's altruism toward HVs) as they differ in nature. 
The sympathy term, $r^{\mathrm{HV}}_{i, k}$ considers the individual reward of the HVs, while the cooperation term, $r^{\mathrm{AV}}_{i, j}$ considers the individual reward of the other AVs, and are defined as
\begin{equation} \label{equ:symreward}
r^{\mathrm{HV}}_{i, k} = \frac{\mathcal{W}_k}{d_{i,k}^\lambda} \sum_m \omega_m x_m \quad 
r^{\mathrm{AV}}_{i, j} = \frac{\mathcal{W}_j}{d_{i,j}^\lambda} \sum_m \omega_m x_m
\end{equation}

in which $d_{i,k}/d_{i,j}$ represents the distance between the agent and the corresponding HV/AV, $\lambda$ is a dimensionless coefficient, $\mathcal{W}_k$ a weight value for individual vehicle's importance, $m$ are the traffic metrics that have been considered in the vehicle's utilities (speed, crashes, etc.), in which $x_m$ is the $m$ metric normalized value and $w_m$ is the weight associated to that metric. The term $r^{\mathrm{M}}$ accounts for the reward of the vehicle's mission. A mission is defined as any desired specific outcome for a particular vehicle, as merging, exiting, etc. 
% The total utility of the mission is defined as a sum of the individual rewards.
%
\begin{equation}
\label{equ:missionreward}
r^{\mathrm{M}}_{i,j} = 
\begin{cases}
\frac{w_j}{(d_{i,j})^\mu}, & \mathrm{if} f(j) \\
0, & \mathrm{o.w.}
\end{cases}
\quad
r^{\mathrm{M}}_{i,k} = 
\begin{cases}
\frac{w_k}{(d_{i,k})^\mu}, & \mathrm{if} f(k) \\
0, & \mathrm{o.w.}
\end{cases}
\end{equation}

The function $f(v)$ is an independent function to evaluate the mission; $f(v)$ return true if the vehicle $v$ has a mission defined and the mission has been accomplished in the recent time window. $\mu$ is a dimensionless coefficient,
$w_j/w_k$ are weights for individual vehicle's mission (importance of the mission).
% and $(d_{i,j})^2/(d_{i,k})^\mu$ account for the relevance of the influence of AV $I_i$ in the mission of vehicle $I_j/V_k$, the higher the distance the lower the relevance thus the reward. 
This allows to define a general reward independent of the driving scenario and mission goals for different vehicles. In our experiments, a \textbf{HV} can be assigned a merging mission or a highway exiting mission, as referred to in Figure~\ref{fig:mainfigure}.

%%%%%%%%%%%%%%%%%%%%%%%%%%%%%%%%%%%%%%%%%%%%%%%%%%%%%%%%%%%%%%%%%%%%%%%

\subsection{Deep MARL architecture for Cooperative Driving}
\begin{figure}[t]
  \centering
  \includegraphics[width=.48\textwidth]{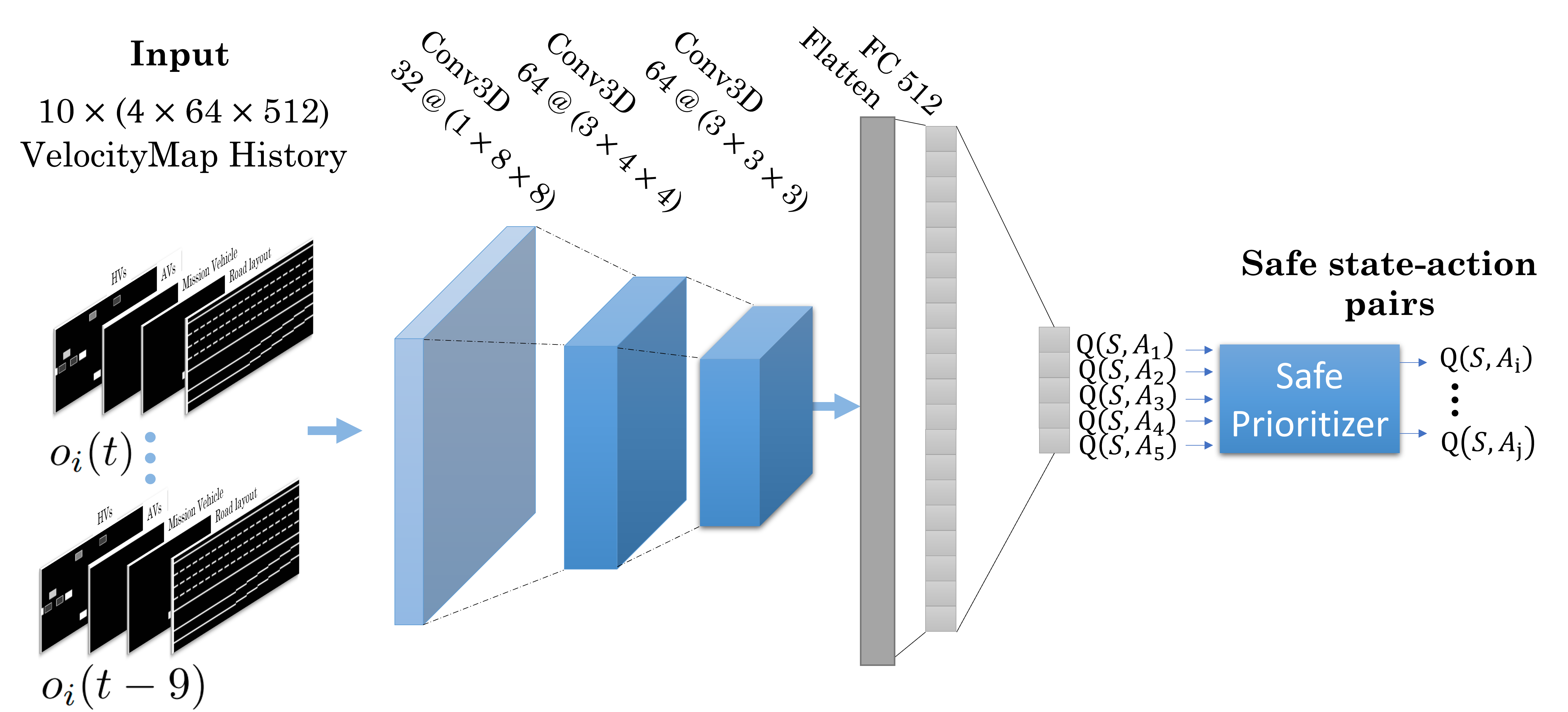}
  \caption{\small{Deep MARL architecture with the safety prioritizer.}}
  \label{fig:architecture}
\end{figure}

We use a 3D Convolutional Neural Network (CNN) with a safety prioritizer as presented in Figure~\ref{fig:architecture}. The 3D CNN acts as a feature extractor and uses a history of VelocityMap observations to account for the temporal information. 

To tackle the non-stationarity of MARL, we train the agents in a semi-sequential approach, as in~\cite{toghi2021social}. The agents are trained independently for $N_{iterations}$ iterations while freezing the policies of the remaining AVs, $\textbf{w}^-$. Subsequently, the other agents' policies are updated with the new policy, $\textbf{w}^+$.
% as explained in Algorithm~\ref{alg:Robust_MARL_algorithm}. 
To improve sample efficiency and train the agent safely, reducing episode resets due to imminent collisions, we use a safety prioritizer that, in the cases where the action selected by the agent policy is unsafe, selects a safe action and stores the unsafe action ($a_t$) and the related state in the $RM$ with a suitable penalty on the reward ($r_{unsafe}$) for the unsafe state-action pair. Those pairs are not removed so the agent can also learn from unsafe experiences. The experience $(\psi(s_{t}), a_t , r_{unsafe} , \emptyset$) is stored in $RM$ with a terminal next state $\emptyset $, the target for this unsafe pair $ (s_t,a_t)$ is $Target(s_t, a_t)^{DDQN} = r_{unsafe}$. 
The details of the safety prioritizer are given in the next section~\ref{sec:safeprioritizer}.

\textbf{Algorithm}~\ref{alg:Robust_MARL_algorithm} summarizes the overall methodology of our safety prioritized deep MARL architecture. Additionally, we do not initiate the learning process until the replay buffer is filled with a minimum number of sample simulations. Moreover, inspired by~\cite{schaul2015prioritized} and~\cite{toghi2021social}, we update our experience replay buffer to compensate for the highly skewed training data. Balancing skewed data is a common practice in machine learning and is beneficial in our MARL problem as well.

% TODO 
%---------------------------------------------------------------------
\begin{algorithm}[t]
    \caption{Safety Prioritized Multi-agent DDQN} 
    \label{alg:Robust_MARL_algorithm}
    % \begin{algorithmic}[1]
    \begin{algorithmic}
         \STATE Initialize \textit{experience replay buffer} $RM$.
         \STATE Initialize $\Tilde{Q}(.;\textbf{w}^-)$ with random weights $\textbf{w}^-=\textbf{w}_{ini}$
         \STATE Initialize target network $\Tilde{Q}(.;\hat{\textbf{w}})$ with weights $\hat{\textbf{w}}=\textbf{w}^-$
          \STATE Pre-store experience of first's 50 episodes in $RM$
        %  \STATE Initialize target action-value function $\Tilde{Q}(.;\hat\textbf{w})$ 
        %  \STATE Initialize $\hat{Q}$ with weights $\hat{\theta}=\theta$.
         \FOR{$\mathrm{e}=50$ to $N_{\mathrm{episode}}$} 
         \STATE Initialize $s_1 =\{ \Tilde{\textbf{o}}_{1}  \}$  and compute $\psi_1 = \psi(s_1)$
              \FOR{t = 1 to T }
                \FOR{$I_i$ in $\mathcal{I}$}
                    \STATE Freeze $\textbf{w}^-$ for all $I_j$, $j \neq i$
                    % \STATE $\textbf{w}^+$ \leftarrow $\textbf{w}^-_i$ 
                     \FOR{$m=1$ to $N_{iterations}$}
                         \STATE With probability $\epsilon$ select a random action $a_t$ ,
                        %  \STATE otherwise select $a_t = \max_{a' \in A} Q(\phi(s_t),a',\textbf{w})$   $\Tilde{\textbf{o}}_{t}$$
                         \STATE otherwise select $a_t = \max_{a' \in A} Q(\psi(s_t),a';\textbf{w}^+)$
                         %TODO update ref to Algorithm 2
                         \IF {$a_t$ is unsafe (\textbf{Algorithm 2})} 
                          \STATE Store $(\psi_t, a_t , r_{unsafe} , \emptyset$) in $RM$
                        %   \STATE Store experience $(\psi(s_t), a_t , r_{unsafe} , \emptyset$) in $RM$
                         \STATE $a_t$ = Compute a safe action (\textbf{Algorithm 3}) 
                            %  \STATE $a_t = a_{safe}$ = Compute a safe action (Algorithm~\ref{alg:safeaction}) 
                        %  \STATE $a_t = a_{safe}$
                         \ENDIF
                          \STATE Execute safe action $a_t$ , and observe $r_t, \Tilde{\textbf{o}}_{t}$ 
                         \STATE Set $s_{t+1} =\{  s_{t},\Tilde{\textbf{o}}_{t+1} \}$ and $\psi_{t+1} = \psi(s_{t+1})$
                         \STATE Store experience $(\psi_t, a_t , r_t , \psi_{t+1}) $ in $RM$
                         \STATE Sample a mini-batch of size $M$ from $RM$
                        %  \STATE Set
                        % 	$y_j =
                        %     \left\{
                        %     \begin{array}{l l}
                        %     %  r_j  \quad & \text{for terminal state} \phi_{j+1}\\
                        %       r_j  \quad & \text{for terminal state}\\
                        %       Target^{DDQN}_j  \quad & \text{otherwise} 
                        %     %   r_j + \gamma \max_{a'} Q(\phi_{j+1}, a'; \theta) \quad & \text{for non-terminal } \phi_{j+1}
                        %     \end{array} \right.$
                        %  \STATE $y_j = r_j$ if episode terminates at step j+1
                        %  \STATE $y_j = r_j + \gamma \max_{a' \in A} \hat{Q}(\phi(s_t),a',\hat{\theta}) $ otherwise
                        %  \STATE Performs a gradient descent step on $(y_j -  Q(\phi_j,a_j,\theta))^2 $ with respect to the network parameters $\theta$
                        % \STATE Compute $\mathcal{L}(\textbf{w}_j)$ ($target_j = y_j $)
                        \STATE Compute $\mathcal{L}(\textbf{w}^+)$ 
                        \STATE Performs gradient descent 
                        \STATE $\textbf{w}^+_{k+1} \leftarrow \textbf{w}^+_k - \alpha \hat{\nabla}_\textbf{w} \mathcal{L}(\textbf{w}^+)$

              \ENDFOR
              \STATE $\textbf{w}^- = \textbf{w}^+$ for all $I_i \in \mathcal{I}$
            \ENDFOR
             %  \STATE Every Target_{steps}  reset $\hat{Q} = Q$
            \STATE Every $\quad Target_{update} \quad$ reset $\hat{\textbf{w}} \leftarrow \textbf{w}^-$
          \ENDFOR
        \ENDFOR
    \end{algorithmic}
\end{algorithm}
%---------------------------------------------------------------------

%%%%%%%%%%%%%%%%%%%%%%%%%%%%%%%%%%%%%%%%%%%%%%%%%%%%%%%%%%%%%%%%%%%%%%%

\subsection{Safety prioritizer}
\label{sec:safeprioritizer}
% Safety
As safety is an essential requirement in autonomous navigation, we add a safety prioritizer to the MARL algorithm, to avoid and penalize imminent collisions. This allows the agent to improve sample efficiency during training and avoid collisions during deployment. If the agent encounters an unseen scenario and decides to take an unsafe action, that action will be avoided. The safety prioritizer improves the simulation results and is critical in real-life situations. The safety prioritizer consists of \textbf{Algorithm 2} and \textbf{Algorithm 3}.

% \noindent
\textbf{Algorithm 2.} During action selection of the agent $I_i$, once an action $a_t$ is chosen, the safety prioritizer checks if the action is safe by computing a safety score for $N_{steps}$ of planning.
We utilize the time-to-collision ($ttc$) as a safety score. If $safety_{score}< safe_{th}$ the action is unsafe and we need to select a safe action. The selection of a safe action is presented in \textbf{Algorithm 3}.

%% TODO Algorithm or text ??
% \noindent
\textbf{Algorithm 3.} The safe action selection is different in training and testing. During training, to encourage exploration, we remove the unsafe actions and keep the random action selection following the current exploration policy on the remaining actions. During testing, we follow the greedy policy in the subset of safe actions $a_t = \max_{a' \in \widetilde{\mathcal{A}}_{safe} } Q(\psi(s_{t}),a';\textbf{w})$. It should be noted that the algorithm does not choose the safest of all possible actions, as that action may lead to particularly conservative behaviors that can compromise traffic efficiency; we instead remove the imminent unsafe actions and follow the priority given by the learned altruistic policy. If it happens that all possible actions are unsafe, we return the action $a_t \in \mathcal{A}$ with the highest safety score. In that way during training the constrained exploration will keep the agent from taking unsafe actions which will lead to sample efficient and more stable learning; and during testing the decision-making is based on the prosocial learned policy with minimum intervention from the safety prioritizer, achieving higher traveled distance while avoiding collisions.

%---------------------------------------------------------------------

%%%%%%%%%%%%%%%%%%%%%%%%%%%%%%%%%%%%%%%%%%%%%%%%%%%%%%%%%%%%%%%%%%%%%%%%%%%%%%%%

\subsection{Modeling Driver Behaviors} % Human Drivers' Decision-making 
\label{sec:humandrivermodel}
%TODO 
We model the longitudinal movements of HVs using the \textit{Intelligent Driver Model} (IDM)~\cite{treiber2000congested}, while the lateral actions of HVs are based on the MOBIL model~\cite{kesting2007general}. The MOBIL model considers two main criteria,

% \noindent 
\textbf{The safety criterion} ensures that after the lane change, the deceleration of the new follower $\mathrm{a}^{}_n$ in the target lane does not exceed a safe limit, i.e, $\mathrm{a}^{}_n>-b_{\mathrm{safe}}$. 

% \noindent 
\textbf{The incentive criterion} determines the advantage of HV after the lane change, quantified by the total acceleration gain, given by
\begin{equation}
\label{equ:mobilcondition}
\mathrm{a}'_{ego}-\mathrm{a}_{ego}+\sin \phi_{ego} \Big( (\mathrm{a}'_n-\mathrm{a}^{}_n) + (\mathrm{a}'_o-\mathrm{a}^{}_o) \Big) > \Delta a_{th}
\end{equation}
where $\mathrm{a}^{}_{o}$, $\mathrm{a}^{}_{n}$ and $\mathrm{a}^{}_{ego}$ represent the acceleration of the original follower in the current lane, the new follower in the target lane and the ego HV, correspondingly, and $\mathrm{a}'_{o}$, $\mathrm{a}'_{n}$, and $\mathrm{a}'_{ego}$ are the equivalent accelerations considering that the ego HV has changed the lane, $\sin \phi_{ego}$ is the politeness factor.
%where $\phi_{ego}$ is the HV's SVO angle.
Finally, the lane change is performed if the safety and incentive criterion are mutually satisfied.

The IDM Model determines the longitudinal acceleration of a HV $\dot{v}_{\mathrm{k}}$ as following,
\begin{equation}
\label{equ:idm1}
\dot{v}_{\mathrm{k}}=\mathrm{a}_\mathrm{max}\Big[ 1- \Big( \frac{v_k}{v_{\mathrm{k}}^0} \Big)^\delta - \Big( \frac{d^*(v_k, \Delta v_k)}{d_k} \Big)^2 \Big]
\end{equation}
in which $v_k$, $d_k$, $\delta$, $\Delta v_k$, $v_{\mathrm{0}}^k$ denote the speed, the actual gap, the acceleration exponent, the approach rate, and the desired speed of the $k^{th}$ HV, respectively. 
% approach rate (velocity difference between the ego-vehicle and the front vehicle)

The desired minimun gap of the $k^{th}$ HV is given by,
\begin{equation}
\label{equ:idm2}
d^*(v_k, \Delta v_k) = d_k^0 +v_kT_\mathrm{k}^0 + \frac{v_k \Delta v_k}{ (2\sqrt{\mathrm{a}_{\mathrm{max}}.\mathrm{a}_{\mathrm{des}}})}
\end{equation}
where $T_k^0$, $d_k^0$, $\mathrm{a}_{\mathrm{max}}$, and $\mathrm{a}_{\mathrm{des}}$ are the safe time gap, the minimum distance, the comfortable maximum acceleration, and deceleration, correspondingly.

The typical parameters for MOBIL model are 
$\sin \phi_{e}=0.5$, $\Delta a_{th} = 0.1 \frac{m}{s^2}$ 
and $b_{\mathrm{safe}} = 4 \frac{m}{s^2}$. Table~\ref{table:idm} shows typically used parameters of the IDM model~\cite{treiber2000congested}.
% %~\cite{treiber2013traffic} or treiber2000congested

% % TODO
\begin{table}[h!]
% \scriptsize
\centering
\caption{\small{Typical parameters for the IDM model}}
\begin{tabular}{ cccccccc } 
% \hline
%                     & \multicolumn{7}{c}{Intelligent Driver Model (IDM)} \\
%  \hline
  \hline
 \textbf{Parameter} & $v^0$ & $T^0$ & $\mathrm{a}_{\mathrm{max}}$ & $\mathrm{a}_{\mathrm{des}}$& $\delta$ & $d^0$  \\
 \hline
 \textbf{Value} & 30 m/s & 1.5 s & 1 m/s$^2$ & 1.5 m/s$^2$ & 4 & 2 m \\
 \hline
\end{tabular}\\[10pt]
\label{table:idm}
\end{table}

% We do not model HVs behaviors we estimate parameters that simulate those behaviors measure by the metrics..
% \noindent 
\textbf{Heterogeneous Driver Behaviors.}
% Aggressive vehicles, is different from egoistic, though there is correlation
% Aggressiveness limit your capacity to care about others , as it increase probability of crashes
% The more aggressiveness in the env, the more necessity for cooperation 
% The discussion around aggressiveness is interesting, actually like humans you can be aggressive and altruistic, but your aggressiveness will limit the scope of your altruism, because natural limits 
Though those parameters are typical used for IDM and MOBIL models, they simulate just one behavior. In order to generate diverse behaviors $\mathcal{B}$, % To test the adaptability of our algorithm it is essential an environment with vehicles that have heterogeneous behaviors, ranging from aggressive to conservative. 
% These behaviors can be modeled using different parameters within a simulator. We 
we frame the task of simulating diverse behaviors as the problem of obtaining the appropriate range of parameters ($\mathcal{P}$) that can generate those behaviors. To achieve that, we leverage a behavior classifier and iteratively simulate the parameters and classify the behaviors, mapping parameters to behaviors. 
%To compute those simulations parameters ($\mathcal{P}$) 
To classify the behaviors we represent traffic using a traffic-graph at each time step $t$, $\mathcal{G}_t$, with a set of edges $\mathcal{E}(t)$ and a set of vertices $\mathcal{V}(t)$ as functions of time, i.e, the positions of vehicles ($ \widetilde{\mathcal{H}} \cup \widetilde{\mathcal{I}} $) represent the vertices. The adjacency matrix $A_t$ is given by $A(k,m) = d(v_k,v_m), k \neq m$ , in which $d(v_k,v_m)$ is the shortest travel distance between vertices $k$ to $m$. Then we use centrality functions~\cite{chandra2020cmetric} to classify the behavior (level of aggressiveness) resulted from $\mathcal{P}$, and then use those simulation parameters $\mathcal{P}$ to model behaviors within the simulator with varying levels of aggressiveness. The centrality functions are defined as,

% \noindent 
\textbf{Closeness Centrality:} the discrete closeness centrality of the $k^\textrm{th}$ vehicle at time $t$ is defined as,
\begin{equation}
    \mathcal{C}^k_C[t] = \frac{{N-1}}{\sum_{v_m\in \mathcal{V}(t)\setminus \{v_k\}} d_t(v_k,v_m)},
    \label{eq: closeness}
\end{equation}
where $N = |\widetilde{\mathcal{H}} \cup \widetilde{\mathcal{I}}|$.
The more central the vehicle is located, the higher $\mathcal{C}^k_C[t]$ and the closer it is to all other vehicles.

% \noindent 
\textbf{Degree Centrality:} the discrete degree centrality of the $k^\textrm{th}$ vehicle at time $t$ is defined as,
\begin{equation}
    \begin{aligned}
    \mathcal{C}^k_D[t] = \bigl | \{ v_m \in \mathcal{N}_k(t) \} \bigr | + \mathcal{C}^k_D[t-1] &\\
    \textrm{such that} \ (v_k,v_m) \not\in \mathcal{E(\tau)}, \tau = 0, \ldots, t-1&
    \end{aligned}
    \label{eq: degree}
\end{equation}

in which $\mathcal{N}_k(t) = \{ v_m \in \mathcal{V}(t), \ A_t(k,m) \neq 0, \nu_m \leq \nu_k\}$ represents the set of vehicles in the proximity of the $k^\textrm{th}$ vehicle, given that $\nu_m \leq \nu_k$; and $\nu_m, \nu_k$ denote the velocities of the $m^\textrm{th}$ and $k^\textrm{th}$ vehicles. The more new vehicles seen by vehicle $k$ that meet this condition, the higher $\mathcal{C}^k_D[t]$.

% We exploit the derivatives and extreme values of these centrality functions to compute the likelihood and intensity of different driving styles.

With the centrality functions we can measure the Style Likelihood Estimate (SLE) for different driver styles~\cite{chandra2020cmetric}. We consider two SLE measures. The SLE of overtaking and sudden lane-changes ($SLE_l$) and the SLE of overspeeding ($SLE_o$).
The $SLE_l$ and $SLE_o$ can be computed by measuring the first derivative of the centrality functions as, 
\begin{equation}
   \textrm{SLE}_l(t) = \abs*{\frac{\partial \mathcal{C}_C(t)}{\partial t}} \quad
   \textrm{SLE}_o(t) = \abs*{\frac{\partial \mathcal{C}_D(t)}{\partial t}}
    \label{eq:sle}
\end{equation}
The maximum likelihood 
$\textrm{SLE}_\textrm{max}$ is calculated as $\textrm{SLE}_{\textrm{max}} = \max_{t \in \Delta t}{\textrm{SLE}}(t)$.

Using those functions, we can approximately quantify and classify driver behaviors in our simulation. The intuition behind that is that an aggressive driver may frequently overspeed or perform sudden lane changes; while overspeeding the $\mathcal{C}_D(t)$ monotonically increases (higher $\textrm{SLE}_o(t)$) and during sudden lane changes the slope and the extrema of $\mathcal{C}_C(t)$ changes values. Thus higher values of $\textrm{SLE}_{\textrm{max}}$ are related to increased levels of aggressiveness. Conversely, conservative drivers are not inclined towards those aggressive maneuvers, and the degree centrality will be relatively flat, 
% the values of the closeness and degree centrality functions in the case of conservative vehicles thus, remain constant, 
thus $ \textrm{SLE}_o(t) \approx 0$ for conservative drivers.

We use these metrics as approximations of the driver's level of aggressiveness. In order to compute the suitable values for our simulation, we iteratively simulate the parameters from IDM and MOBIL models, and for each set of parameters, we quantify the resulting behavior in the simulation (using those metrics). Mapping the parameters $\mathcal{P}$ to behaviors (quantified in the simulation for those parameters).
The estimated simulation parameters that simulate conservative, moderate and aggressive behavior in our scenarios are presented in Table~\ref{table:parameters}.

% estimated simulation parameters that define conservative, moderate and aggressive behavior in our scenarios.
% Finally we get the following parameters , and for other behaviors generate values in between then that will go from aggressive to conservative.

\begin{table}[h]
\caption{\small{Estimated simulation parameters that simulate conservative, moderate and aggressive behavior in our scenarios.}}
\centering
\begin{tabular}{c|cccc} 
\hline

Model & Parameter & Aggressive  &  Moderate & Conservative \\
\hline
\hline
MOBIL & $\sin \phi_{e}$ & 0  & 0.3 & 1\\ % MOBIL& Politeness ($\sin \phi_{e}$)
& $\Delta a_{th}$ & 0 $m/s^2$ & 0.1 $m/s^2$ & 0.4 $m/s^2$ \\ % Acceleration gain ($\Delta a_{th}$)
& $b_{\mathrm{safe}}$ & 12.0 $m/s^2$ & 6.0 $m/s^2$ & 2.0 $m/s^2$\\ % Safe limit ($b_{\mathrm{safe}}$)

\hline
IDM & $T^0$ & 0.5s   & 1s & 3s \\ % Time gap ( $T^0$)
& $d^0$ & 1 $m$ & 2 $m$ & 6.0 $m$\\ % Minimum distance ($d^0$ )
& $\mathrm{acc}_{\mathrm{max}}$  & 7.0 $m/s^2$ & 3.0 $m/s^2$ & 1.0 $m/s^2$\\ % Maximum acceleration. ($\mathrm{acc}_{\mathrm{max}}$) 
& $\mathrm{acc}_{\mathrm{des}}$ & 12.0 $m/s^2$        & 7.0 $m/s^2$ & 2.0 $m/s^2$\\ % Maximum deceleration ($\mathrm{acc}_{\mathrm{des}}$)
\hline
% \hline
\end{tabular}

\label{table:parameters}
\vspace{-10pt}
\end{table}

The desired velocity $v^0$ is set to $30m/s$ and the acceleration exponent $\delta = 4$.

%%%%%%%%%%%%%%%%%%%%%%%%%%%%%%%%%%%%%%%%%%%%%%%%%%%%%%%%%%%%%%%%%%%%%%%%%%%%%%%%
%%%%%%%%%%%%%%%%%%%%%%%%%%%%%%%%%%%%%%%%%%%%%%%%%%%%%%%%%%%%%%%%%%%%%%%%%%%%%%%%

%%%%%%%%%%%%%%%%%%%%%%%%%%%%%%%%%%%%%%%%%%%%%%%%%%%%%%%%%%%%%%%%%%%%%%%%%%%%%%%%

\subsection{Computational Details and Hyperparameter} % Implementation and Hyperparameter Tuning
\vspace{-0.1cm}
We customize the OpenAI Gym environment in~\cite{leurent2019approximate} to suit our particular driving scenario and MARL problem.
The PyTorch implementation of our architecture on average takes 3.1GB of memory for 4 agents and 18 HVs. Using a GPU NVIDIA Tesla V100. The training process is repeated several times to ensure convergence of the experiments to a similar policy. The network is trained for $N_{episodes} = 10,000$ taking on average 8 hours and a forward pass during testing requires on average 15ms. We utilize 3,200 GPU-hours for our simulations. Table~\ref{table: hyperparameters} lists our simulation and training hyper-parameters. 

%---------------------------------------------------------------------
\vspace{10pt}
\begin{table}[t]
\caption{\small{Simulation and training hyper-parameters.}}
\begin{center}
\begin{tabular}{c c | c c}
\hline
Parameter &
Value &
Parameter &
Value\\ 
\hline
\hline
$N_{\mathrm{episode}}$ &
10,000 &
$\epsilon$ decay &
Linear  \\
$RM$ buffer size &
8,000 &
Initial exploration $\epsilon_0$ &
1.0 \\ 
Batch size & 
32 &

Final exploration & %$\epsilon_f$
0.05 \\ 
Learning rate $\alpha_0$ &
0.0005 &
Optimizer &
ADAM \\ 
$Target_{update}$ &
300 &
Discount factor $\gamma$ &
0.95 \\ 
$|\mathcal{H}|$&
$18$ &
$|\mathcal{I}|$ &
$4$ \\ 
% $T_{\mathrm{episode}}$ &
% $18s$ &
\hline
% \hline
\end{tabular}
\end{center}
\label{table: hyperparameters}
\end{table}
%---------------------------------------------------------------------

\section{Experimental Results} 
\label{sec:experiments}

%%%%%%%%%%%%%%%%%%%%%%%%%%%%%%%%%%%%%%%%%%%%%%%%%%%%%%%%%%%%%%%%%%%%%%%
% \subsection{Controlled Variables} % variables
% \label{sec:controlledvariables}

\noindent
% explain variables better ,safe_{th} , level of aggressiveness or driver behaviors, 
\textbf{Controlled Variables.}
% 1)---Controlled/Manipulated variables
We study how the $safe_{th}$, the \emph{level of aggressiveness}, the \emph{traffic scenarios} ($f_j$) and the \emph{HVs' behaviors} ($b_k$) impact the performance of AVs. We consider the case in which the mission vehicle (merging/exiting) in Figure~\ref{fig:mainfigure} is \emph{human-driven}, $M \in \mathcal{H}$, and define the following terms: 
% \textbf{$AV_S$/$AV_E$} as safer social/egositic AVs that act \emph{altruistically/egoistic} in the presence of diverse HVs behaviors $b \in \mathcal{B}$ .
\begin{itemize}
    % \item \textbf{$AV_S$/$AV_E$}. Safer social/egositic AVs that act \emph{altruistically/egoistic} in the presence of diverse HVs behaviors $b \in \mathcal{B}$ .
    \item \textbf{$AV_S$}. Social AV ($\phi_i = \phi^*$) that act \emph{altruistically} in the presence of diverse HVs behaviors $b \in \mathcal{B}$.
    \item \textbf{$AV_E$}. Egoistic AV ($\phi_i = 0$) that act \emph{egoistically} in the presence of diverse HVs behaviors $b \in \mathcal{B}$.
%     \item \textbf{Adaptation}. Safer Social AVs act \emph{altruistically} and are trained in different scenarios $f_i \in \mathcal{F}$ and tested in other scenarios $f_j \in \mathcal{F} , f_j \neq f_i$.
%     \item \textbf{$AV_S$} AVs act \emph{altruistically} without a safety prioritizer
%     % \item \textbf{$AV_S$} (Safer Social AV). Autonomous agents act \emph{pro-socially} with a safety prioritizer
%     % \item \textbf{Social AV. } The mission vehicle is \emph{human-driven} and autonomous act pro-socially,
    
%     % \emph{autonomous}.
\end{itemize}

% In our experiments we use the optimal SVO angle $\phi^*$  as in~\cite{toghi2021social}.
with $\phi^*$ to be the optimal SVO angle tuned to reach the optimal level of altruism as in~\cite{toghi2021social}.
% For safer social AVs, as in~\cite{toghi2021social} we tune SVO angle to reach the optimal level of altruism with $\phi^*$ been the optimal SVO angle. In~\ref{equ:decentralizedreward} SVO can be varied from $\phi=0$ (purely egoistic) towards $\phi=\pi/2$ (purely altruistic). However neither of the two extremes are optimal and a point between leads to the most socially desirable outcome, which is the optimal SVO angle $\phi^*$. Therefore, in our experiments we use the optimal SVO angle $\phi^*$ of when training $AV_S$ agents.
%%%%%%%%%%%%%%%%%%%%%%%%%%%%%%%%%%%%%%%%%%%%%%%%%%%%%%%%%%%%%%%%%%%%%%%
% \vspace{-0.1cm} 
% \subsection{Performance Metrics} % measures
% \label{sec:performancemeasures}

% \noindent
\textbf{Performance Metrics.}
% 1)---Performance Metrics/Measures
We measure the performance of our system based on safety, efficiency, altruistic performance gain ($PG$) and adaptation error $\mathrm{A}_\mathrm{error}$. To measure safety, we compute the percentage of episodes that encountered a crash ($C(\%)$). For efficiency, the average traveled distance ($DT(m)$) of the vehicles and the number of missions accomplished by the mission vehicle are used. The altruistic performance gain is measured by computing the difference in the safety/efficiency performance of \textbf{$AV_E$} and \textbf{$AV_S$}, as
\begin{equation}
    PG_{safety}(\%) = \frac{(AV_E)_{C(\%)} - (AV_S)_{C(\%)}}{N_{Episodes}} 
\end{equation}

\begin{equation}
    PG_{efficiency}(\%) = \frac{(AV_S)_{DT(m)} - (AV_E)_{DT(m)}}{(AV_E)_{DT(m)}} 
\end{equation}
% and efficiency performance gain  $ PG_{eff} = \frac{(AV_E)_{DT} - (AV_S)_{DT}}{Max_{distance}}$ with $Max_{distance} = Max_{speed}*Simulation_{tim}$ .
Finally the adaptation error is a weighted sum function of the safety ($C(\%)$) and efficiency ($DT(m)$) performance of the \textbf{$AV_S$} when trained and tested in different scenarios/behaviors. Defined as,
\begin{equation}
    A_{error}(\%) = w_{s}\times (C(\%)) + w_{e}\times 100(1-\frac{DT}{DT_{max}})
\end{equation}
such that an adaptation between different situations that result in $0\%$ crash and $DT = DT_{max}$ will have $\mathrm{A}_\mathrm{error}=0\%$. 
%Using those metrics, we can measure the safety, efficiency, robustness and adaptability of our system.
% Maximum distance max speed*simulation time
% safety, number of crashes
% efficiency, distance traveled and missions accomplished 

% \vspace{-0.1cm}
% \subsection{Dependent Measures}
%%%%%%%%%%%%%%%%%%%%%%%%%%%%%%%%%%%%%%%%%%%%%%%%%%%%%%%%%%%%%%%%%%%%%%%
\vspace{-0.2cm}
\subsection{Hypotheses} % connect variables to measures
% 1)---Connect manipulated variables and performance measures
% we study the impact of cooperation to driver behaviors. We show how prosocial behaviors are important and their sensitivity vs different behaviors , the more aggressiveness the more important cooperation is
%% TODO We open the discussion  aggressiveness and egoistic ???
% we present the situations in which a decentralize MARL can adapt to driver behaviors, and study/give insights on how leverage adaptation and transfer learning between traffic scenarios and driver behaviors

In this section we examine the following hypotheses 
%% TODO  should we present it as questions that we attempt to answer ??
% provide experimental results that answer the hypotheses 
\begin{itemize}
    \item \textbf{H1.} \emph{The higher the level of aggressiveness in a mixed-autonomy scenario, the greater the impact of cooperation. Thus, we expect a higher performance gain ($PG$) when altruistic AVs face environments with higher level of aggressiveness}.
    % TODO Connect variable and measure better
    \item \textbf{H2.} \emph{Altruistic AVs agents using the decentralized framework can adapt to different driver behaviors and traffic scenarios without compromising the overall traffic metrics. However, the higher the similarity of testing scenarios to the ones seen during training ($(f_{test},b_{test}) \approx (f_{train},b_{train})$), the lowest adaption error ($\mathrm{A}_\mathrm{error}$)}.
    % We anticipate a higher adaptation error in similar scenarios, and expect adaptation to be better from challenging to less challenging scenario and not the other way}.
    \item \textbf{H3.} \emph{With the inclusion of the safety prioritizer, we anticipate improvement in safety and efficiency. We expect that AVs will cause more crashes in the absence of a safety prioritizer ($safe_{th}=0$)}.
    % \item \textbf{H4.} \emph{The use of transfer learning expedite the speed of convergence as an efficient methods to transfer knowledge from simple concepts to hard problems; however, we hypothesize that not necessary will improve the final model performance of RL agents.}

%?? are manipulated variables and performance measures connected clearly ???
\end{itemize}

%%%%%%%%%%%%%%%%%%%%%%%%%%%%%%%%%%%%%%%%%%%%%%%%%%%%%%%%%%%%%%%%%%%%%%%

\subsection{Analysis and Results}
\label{sec:results}
 Based on the hypotheses, we explore their correctness through the experiments in this section.

%%%%%%%%%%%%%%%%%%%%%%%%%%%%%%%%%%%%%%%%%%%%%%%%%%%%%%%%%%%%%%%%%%%%%%%
\subsubsection{Sensitivity analyses to HVs behaviors} %Sensitivity analyses (altruistic gain)
\begin{figure}[t]
  \centering
  \includegraphics[width=.48\textwidth]{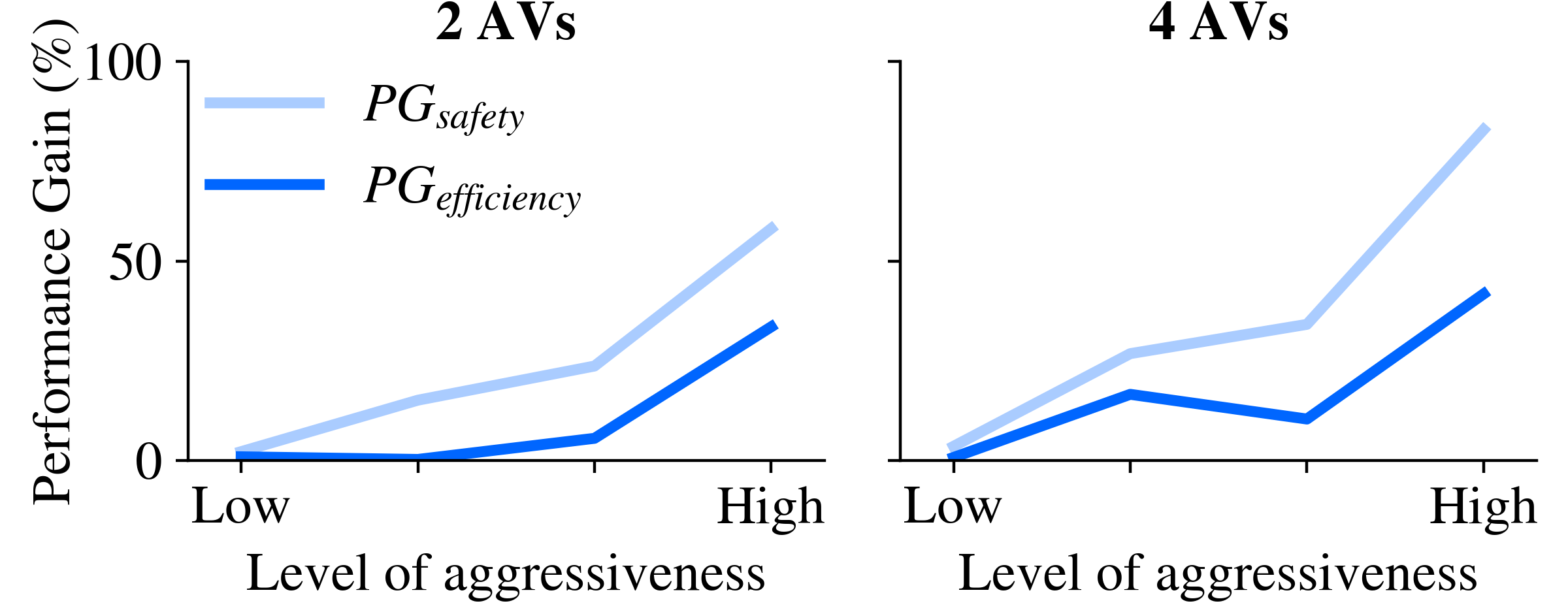}
  \caption{\small{Sensitivity analyses measured by altruistic performance gain (PG) of AVs, the more aggressiveness of the HVs, the higher the impact/gain of cooperation.}}
  \label{fig:sensitivity1D}
\end{figure}

\begin{figure}[t]
  \centering
  \includegraphics[width=.48\textwidth]{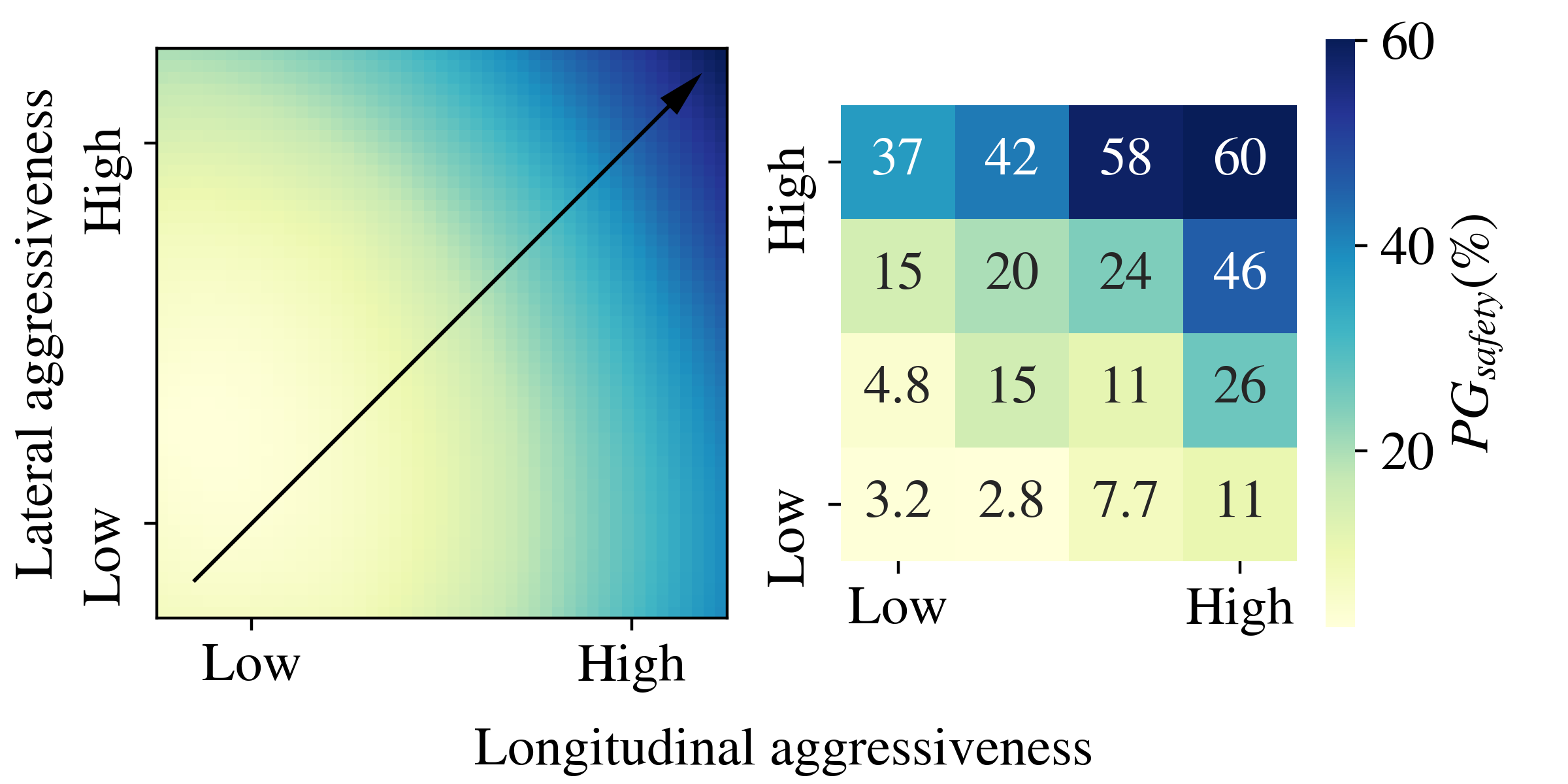}
  \caption{\small{Lateral and longitudinal sensitivity analyses, the altruistic performance gain (PG) increase in both lateral and longitudinal directions. 
  }}
  \label{fig:sensitivity2D}
\end{figure}

To study the hypothesis \textbf{H1} we investigate the effect of HV behaviors on the altruistic AV agents. We focus on scenarios with a HV mission vehicle, with safe AVs that act \emph{altruistically} ($AV_S$) or \emph{egoistic} ($AV_E$) , in environments with increasing levels of HVs aggressiveness. 
% We run simulations with increasing levels of aggressiveness for $AV_S$ and $AV_E$ agents and compare their performance by measuring the altruistic performance gain. 
Figure~\ref{fig:sensitivity1D} illustrates the altruistic performance gain for increasing levels of HVs' aggressiveness for 2 AVs (left) and 4 AVs (right). It demonstrates that the more aggressive the HVs are, the higher is the impact of cooperation and thus confirms the \textbf{H1}.
%It shows the impact of cooperation when the HVs aggressiveness increase with higher gains from cooperative altruistic AVs when the level of aggressiveness is higher, and confirms our \textbf{H1}. Agents that consider social elements ($AV_S$) in their reward show superior performance compared to egoistic agents($AV_E$) and this performance is very sensitive to the level of aggressiveness.
This is also observed in Figure~\ref{fig:sensitivity2D} where the level of aggressiveness is decomposed into lateral and longitudinal aggressiveness. Lateral and longitudinal aggressiveness is varied by changing the MOBIL and IDM parameters (Table~\ref{table:parameters}) from aggressive to conservative. Figure~\ref{fig:sensitivity2D} shows that the altruistic gain increases in both directions, but is more pronounced in the longitudinal direction. That is probably due to the simulated scenarios having more longitudinal maneuvers. 
% We conclude that the higher the traffic aggressiveness, the higher the importance of cooperation.
 
%---------------------------------------------------------------------
%---------------------------------------------------------------------

%---------------------------------------------------------------------
%---------------------------------------------------------------------

%%%%%%%%%%%%%%%%%%%%%%%%%%%%%%%%%%%%%%%%%%%%%%%%%%%%%%%%%%%%%%%%%%%%%%%
% TODO clean the text , no clear yet
\subsubsection{Domain adaptation of altruistic agents}
\label{sec:deepnetworks}
\begin{figure}[t]
  \centering
  \includegraphics[width=.48\textwidth]{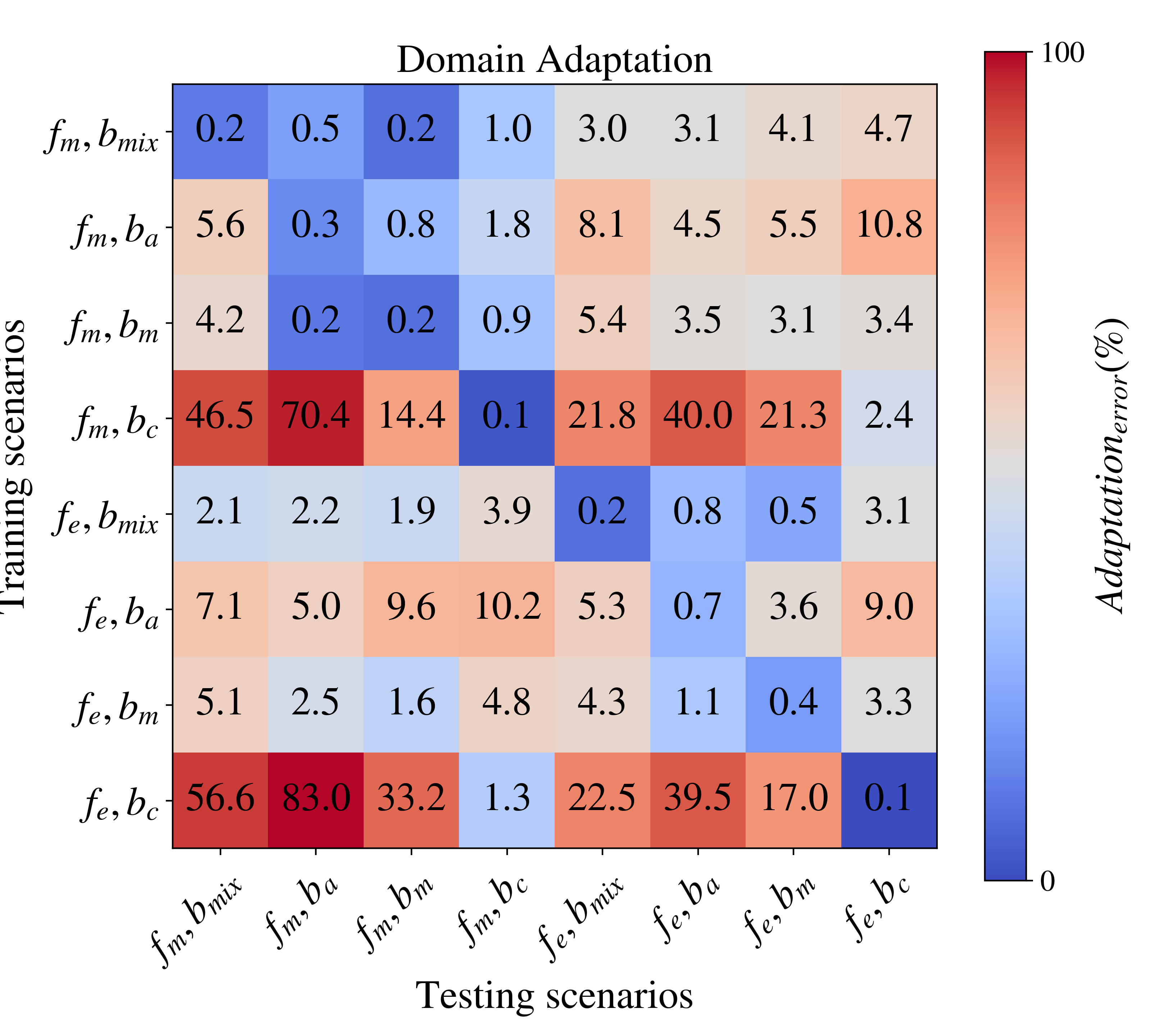}
  \caption{\small{The domain adaptation matrix with adaptation error ($\mathrm{A}_\mathrm{error}$) between different traffic scenarios and behaviors.  $\mathrm{AV}_\mathrm{S}$ are trained (rows of the matrix) in different scenarios $f_i \in \mathcal{F}$ in the presence of HVs with different behaviors $b_k \in \mathcal{B}$ and tested (columns of the matrix) in other scenarios $f_j \in \mathcal{F}$ and behaviors $b_l \in \mathcal{B}$. Each pair ($f_i,b_k$) is a combination of scenario and behavior.
  The lower $\mathrm{A}_\mathrm{error}$ the most suitable the adaptability between those domains.}}
  \label{fig:adaptition}
\end{figure}

% Adaptation
Following the sensitivity analysis, we investigate the domain adaptation of the AVs to validate the \textbf{H2}. Figure~\ref{fig:adaptition} shows how the altruistic AVs learn to adapt to different scenarios and behaviors, based on an adaptation score.  
% Fig~\ref{fig:adaptition} presents the result of the adaptation error.
For the experiments, $AV_S$ are trained in different scenarios $f_i \in \mathcal{F}$ in the presence of HVs with different behaviors $b_k \in \mathcal{B}$ and tested in other scenarios $f_j \in \mathcal{F}$ and behaviors $b_l \in \mathcal{B}$. In our experiments, we consider two case study scenarios $f_m , f_e \in \mathcal{F} $ (merging/exiting) in environments with three different HVs behaviors $b_a, b_m, b_c \in \mathcal{B}$ (aggressive, moderate, conservative) see Table~\ref{table:parameters}; and a mixed behavior environment, in which HVs are created randomly and their behaviors are selected based on a uniform distribution over the behaviors in $\mathcal{B}$, given equal probability to the defined behaviors. In total, we have eight combinations of scenarios and behaviors, namely: ($f_m,b_{mix}$), ($f_m,b_a$), ($f_m,b_m$), ($f_m,b_c$), ($f_e,b_{mix}$), ($f_e,b_a$), ($f_e,b_m$), ($f_e,b_c$). 

\begin{figure}[t]
  \centering
  \includegraphics[width=.48\textwidth]{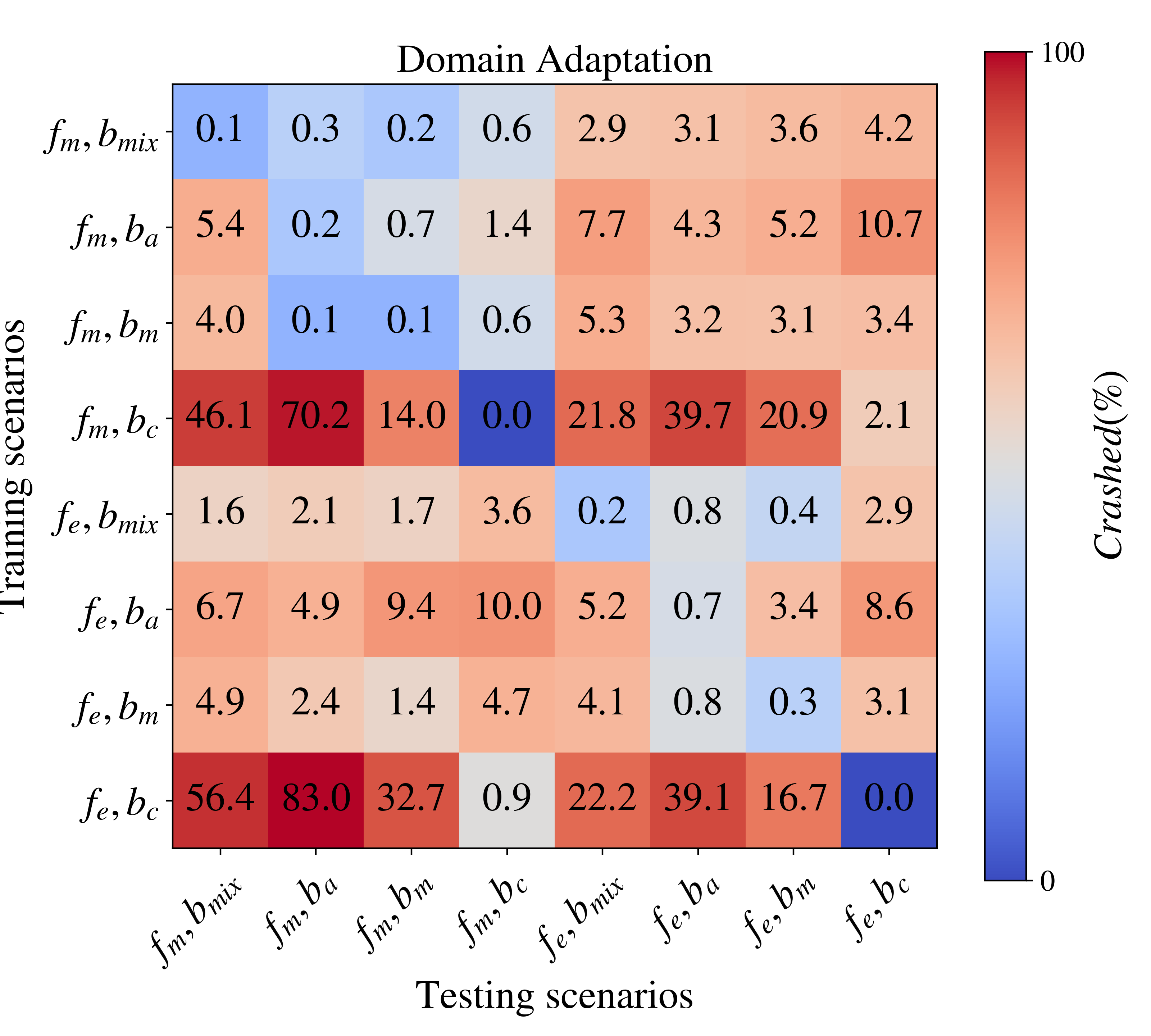}
  \caption{\small{The domain adaptation matrix with crash percentage ($C(\%)$) between different traffic scenarios and behaviors.The lower $C(\%)$ the most suitable the adaptability in terns of safety (measured by $C(\%)$) between those domains. }}
  \label{fig:adaptition_crash}
\end{figure}

\begin{figure}[t]
  \centering
  \includegraphics[width=.48\textwidth]{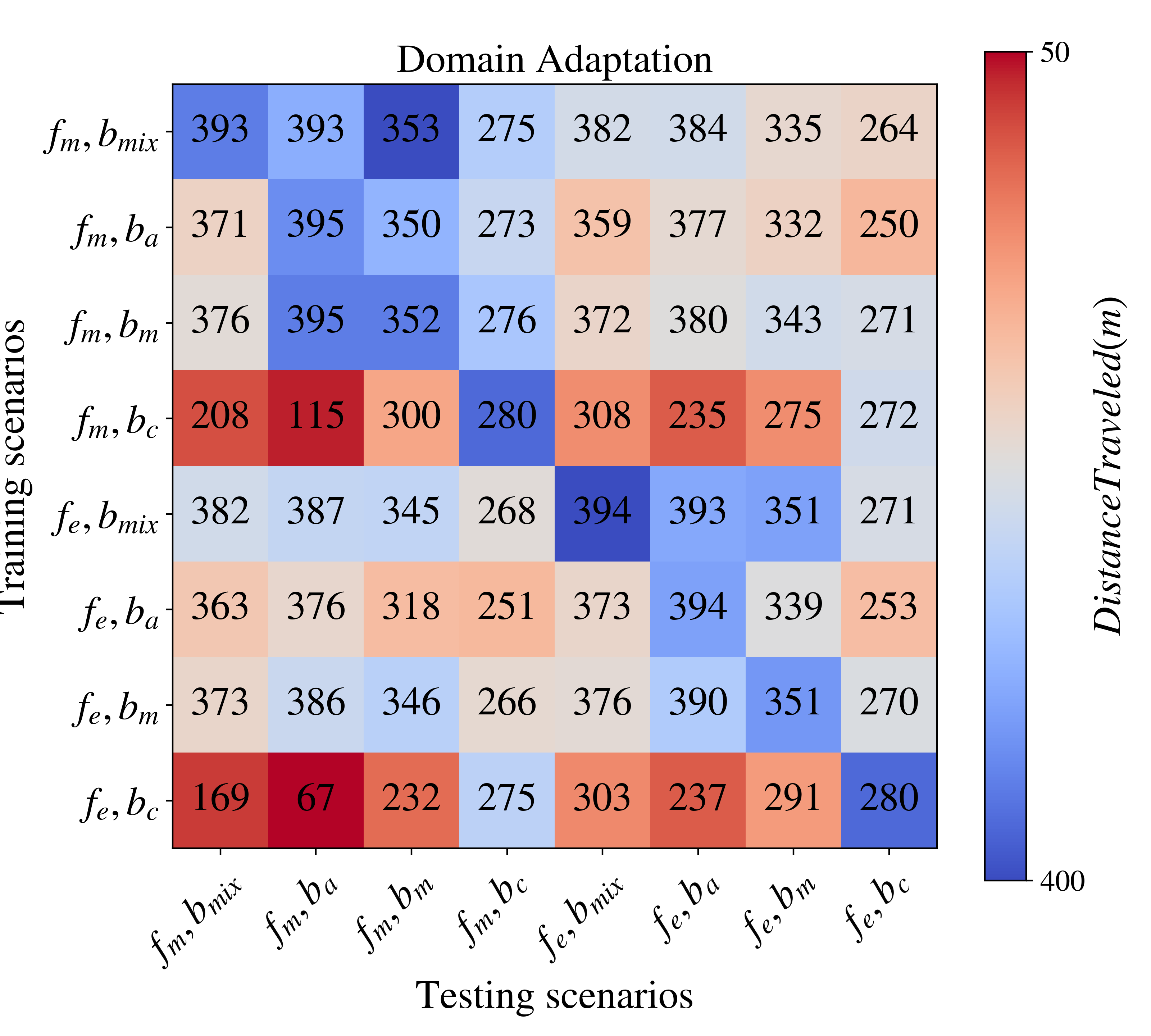}
  \caption{\small{The domain adaptation matrix with distance traveled ($DT(m)$). Illustrating how the AVs adapt to other situations in terms efficiency (measured by $DT(m)$).}}
  \label{fig:adaptition_distance}
\end{figure}

The results are presented in Figure~\ref{fig:adaptition} as an adaptation matrix, showing the $\mathrm{A}_\mathrm{error}$ for different domains, the $\mathrm{A}_\mathrm{error}$ is in percentage ($\%$) and color-map in logarithmic scale to increase the perceived dynamic range for visualization. In our analyses, the weights used for $\mathrm{A}_\mathrm{error}(\%)$ are $w_{s} = \frac{2}{3}$ and $w_{e} = \frac{1}{3}$, which weighs the safety performance higher. $DT_{max}$ is computed based on the maximum distance for each situation. 
% DT is normalized based on $DT_{max}$ for that situation... to converted in % to be added to adaptation score....
Additionally, Figure~\ref{fig:adaptition_crash} and Figure~\ref{fig:adaptition_distance} illustrate how the AVs adapt in terms of safety (measured by $C(\%)$) and efficiency (measured by $DT(m)$), separately.

The matrix shows the best performances in the diagonal as agents trained and tested in the same environment (($f_i,b_k$); ($f_j,b_l$) with $i = j$ and $k = l$) experience during testing similar situations to the ones seen in training. The vehicles trained in the merging environment are able to perform the exiting mission for different behaviors, and vice-versa. It is interesting to notice that when trained in a conservative environment ($b_c$), the performance when tested in aggressive environments ($b_a$) is poor. 
We believe that the reason is that in conservative environments, the HVs yield the mission vehicle, and the AVs learn to rely on HVs to guide the traffic. This learned policy is valid in a conservative environment where one can expect the HVs to always create a safe space for the mission vehicle. However, the same is not valid in more aggressive environments, in which AVs have to guide the traffic to avoid dangerous situations. As a result, the performance of vehicles trained in a conservative environment and tested in an aggressive one is the worse.

% On the other hand, agents trained in an aggressive environment and tested in a conservative environment performs better that their counterpart, though still multiple crashes happened. 
On the other hand, an adequate performance adaptation (lower $\mathrm{A}_\mathrm{error}$) is obtained when agents are trained in the presence of all moderate HVs ($b_m$) or a mixed behavior environment ($b_{mix}$), in which AVs face situations where the HVs yield, but also situations that require learning how to guide the traffic to optimize for the social utility. The results from the domain adaptation matrix indicate that a moderate or mixed environment is the most suitable for training robust AVs and show the adaptability of AVs to different situations, thereby confirming the \textbf{H2} hypothesis.

It can be concluded that the adaptation between the environments is not reciprocal and the selection of the environment and situations should be considered during training, based on the application needs and target situations. The adaptation matrices serve as reference and provide insights on domain adaptation in mixed-autonomy traffic, the matrices present the settings in which altruistic AVs can best learn cooperative policies that are robust to different traffic scenarios and human behaviors. 
% The adaptation matrices serve as reference and suggest settings for learning,
%---------------------------------------------------------------------
%---------------------------------------------------------------------
%---------------------------------------------------------------------
%---------------------------------------------------------------------
%---------------------------------------------------------------------
%---------------------------------------------------------------------

% \noindent\textbf{Transfer Learning} 
\subsubsection{Transfer Learning}
\begin{figure}[t]
  \centering
  \includegraphics[width=.48\textwidth]{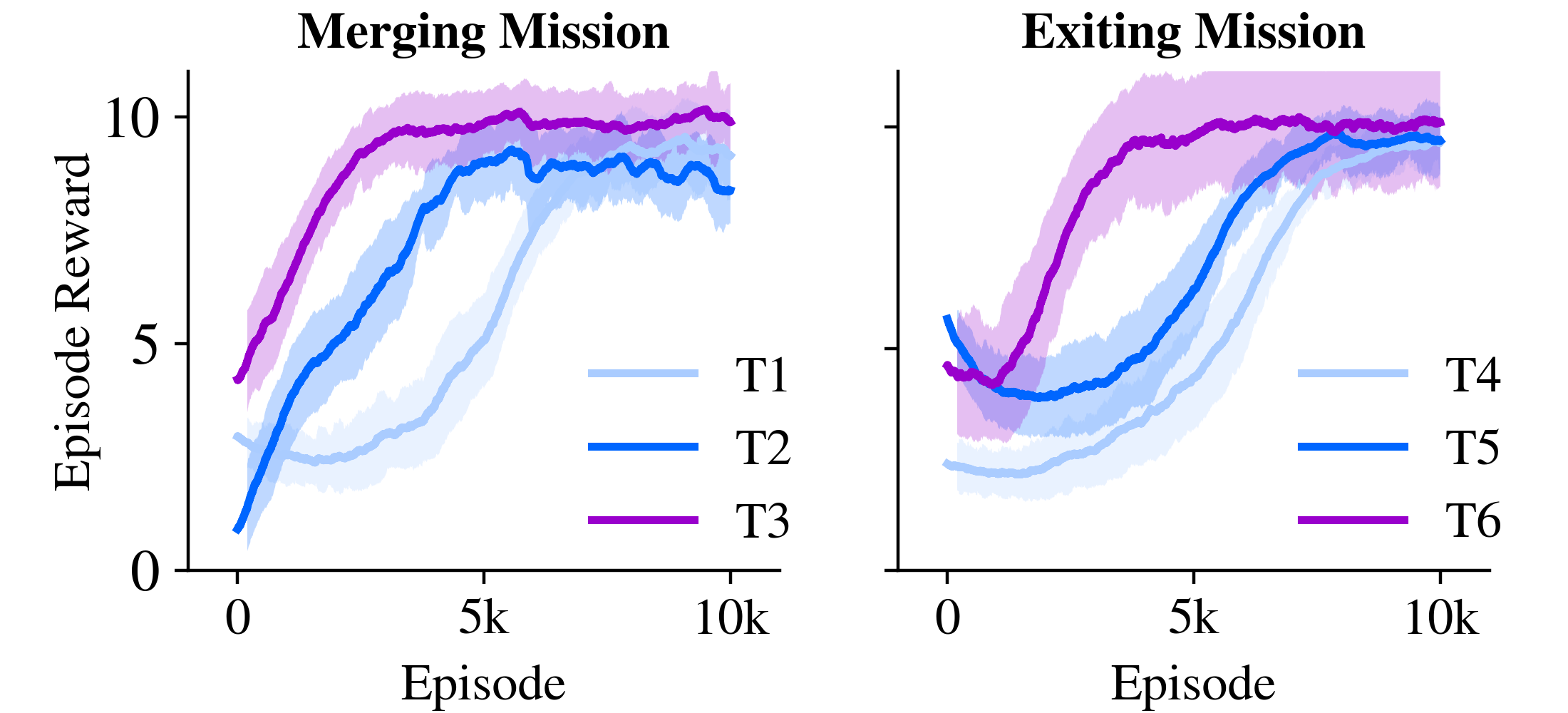}
    \caption{\small{Transfer learning performance. Showing how policies learned during merging can be transferred to the exiting environment to speed up the learning process while archiving similar performance as when learning the task from scratch.}}
%   \caption{\small{Transfer Learning performance. T1:$AV_{merging}$, T2:$AV_{drive-to-merging}$, T3:$AV_{exiting-to-merging}$, T4:$AV_{exiting}$ T5:$AV_{drive-to-exiting}$, T6:$AV_{merging-to-exiting}$}}
  \label{fig:transferlearning}
\end{figure}

Together with domain adaptation we exploit transfer learning to foster generalization while efficiently learning harder tasks from trained models and therefore accelerate the learning. We study how the policies learned during merging can be transferred to the exiting environment. For that, we train AVs agents from scratch for the mission/task of merging $\mathrm{AV}_\mathrm{merging}$ (T1), train AVs agents to drive on a highway, and then use that model as the starting point to learn the merging task $\mathrm{AV}_\mathrm{drive-to-merging}$ (T2), train AVs agents for the exiting task and then use that model as the starting point to learn the merging task $\mathrm{AV}_\mathrm{exiting-to-merging}$ (T3); and apply the same procedure for the exiting task, learning to exit from scratch $\mathrm{AV}_\mathrm{exiting}$ (T4), after learned how to drive $\mathrm{AV}_\mathrm{drive-to-exiting}$ (T5) and after learned how to merge $\mathrm{AV}_\mathrm{merging-to-exiting}$ (T6). The results of the experiments are presented in Figure~\ref{fig:transferlearning} and show that our transfer learning approach speeds up the learning process while archiving similar performance as when learning the task from scratch. 

%---------------------------------------------------------------------
%---------------------------------------------------------------------

%~\cite{mnih2013playing}

\subsubsection{Safety}
% compare with safety prioritizer and without
Finally, we compared state of the art architectures related to our approach~\cite{van2016deep,toghi2021social,toghi2021cooperativearxiv,toghi2021altruisticarxiv} in terms of safety and efficiency to validate \textbf{H3}. We trained the different architectures in the same situations and examined their performance under different levels of HVs behaviors. As noted in Table~\ref{table:comp} our safe altruistic agents consistently outperformed the other approaches, and the results are more notable when the level of aggressiveness is higher. We conclude that when using the safety prioritizer, immediate collisions are avoided reducing the overall number of crashes in the episodes.

%---------------------------------------------------------------------
% **TODO change tables format to look nicer
% \begin{table}[h!]
% \vspace{-5pt}
\begin{table*}[t]
\caption{\small{Architectures' performance comparison. Our safe altruistic AVs outperformed the others approaches.}}
\begin{center}
% \begin{tabularx}{0.95\textwidth}{|X  X X|X X|X X|X X|}
\begin{tabular}{c| c c c| c c c | c c c}
% \hline
\hline
&
\multicolumn{3}{c}{Aggressive HVs} &
\multicolumn{3}{c}{Moderate HVs} &
\multicolumn{3}{c}{Conservative HVs}
\\
 \hline 
 \hline
Approaches &
C (\%)&
MF (\%) &
DT (m)&
C (\%)&
MF (\%) &
DT (m)&
C (\%)&
MF (\%) &
DT (m)\\
\hline

Conv2D+DQN~\cite{van2016deep}& 31.2 &
28.9 &
316 &
25.4 &
20.3 &
302 &
14.0&
7.9 &
274 \\
% \hline

Toghi~\etal~\cite{toghi2021social}&
21.3 &
16.4 &
339 &
12.7&
10.1 &
333 &
1.6 &
0.6 &
269 \\
% \hline

Conv3D+A2C~\cite{toghi2021altruisticarxiv} &
14.8 &
12.6 &
341 &
9.4 &
8.8 &
328 &
1.1 &
0.1 &
267 \\
% \hline

Conv3D+DQN~\cite{toghi2021cooperativearxiv} &
3.1 &
2.8 &
359 &
2.6 &
2.4 &
341 &
0.3 &
\textbf{0} &
\textbf{284} \\
% \hline
% & 
% & 
% & 
% & 
% & 
% & 
% & 
% & 
% & 
% \\
\textcolor{black}{\textbf{Ours}} & 
\textcolor{black}{\textbf{0.2}} & 
\textcolor{black}{\textbf{0.1}} & 
\textcolor{black}{\textbf{397}}& 
\textcolor{black}{\textbf{0.1}} & 
\textcolor{black}{\textbf{0.1}} & 
\textcolor{black}{\textbf{354}} & 
\textcolor{black}{\textbf{0}} & 
\textcolor{black}{\textbf{0}} & 
\textcolor{black}281\\
\hline
% \hline
% \end{tabularx}
\end{tabular}
\end{center}
\raggedright\footnotesize{\hspace{0.4cm} \quad \quad \quad \quad \quad  C: \emph{Crashed}, MF: \emph{Mission Failed}, DT: \emph{Distance Traveled}}\\
\label{table:comp}
% \end{table}
\end{table*}
% \footnote{footnote text}
% \vspace{-5pt}
%---------------------------------------------------------------------

%%%%%%%%%%%%%%%%%%%%%%%%%%%%%%%%%%%%%%%%%%%%%%%%%%%%%%%%%%%%%%%%%%%%%%%%%%%%%%%%
\section{Conclusion and Future Work}
\label{sec:concluding}
% 	State main findings: Emphasize your main results.
%---------------------------------------------------------------------

We study the problem of multi-agent maneuver-level decision-making in mixed-autonomy environments and investigate how AVs can learn cooperative policies that are robust to different scenarios and driver behaviors safely. Our altruistic AVs learn the decision-making process from experience, considering the interests of all vehicles while prioritizing safety and optimizing a general decentralized social utility function. We expose the settings for our MARL problem in which transfer learning and domain adaptation are more feasible, and conducted a sensitivity analysis under different HVs' behaviors.
%Our results demonstrate superior performance in competitive driving scenarios outperforming state-of-the-art works and improving traffic flow, safety, and efficiency. 
Our safe altruistic AVs learn to coordinate and influence the behavior of HVs with socially advantageous results in diverse situations. 

%---------------------------------------------------------------------

%---------------------------------------------------------------------
\smallskip
% \noindent 
\textbf{Limitations and Future Work. } % State any limitations
While we explored different aspects of social navigation in various environments and in the presence of diverse HVs behaviors, the HV models are not learned from real human drivers’ data and the traffic scenarios are limited to merging and exiting. Nevertheless, we speculate that our approach could be effective in realistic traffic situations by utilizing and learning from real human data and traffic scenarios.
Additionally, extra emphasis is required on safety for this approach to be utilized in the real-world scenarios.

%% TODO  Future work  Make recommendations?, Predictions for future developments
In future work, we plan to investigate more sophisticated architecture and state representations, as well as develop a more realistic simulation environment that incorporates data from real-world traffic and can handle more complex interactions between HVs and AVs and diverse traffic agents such as bicycles or pedestrians.
Despite the limitations, we are thrilled to see safe and robust social AVs on the road that learn from experience. We also anticipate applications of these ideas beyond driving, to general MA humans-robot interactions in which agents influence humans and cooperate safely for a socially advantageous outcome.

%---------------------------------------------------------------------

%%%%%%%%%%%%%%%%%%%%%%%%%%%%%%%%%%%%%%%%%%%%%%%%%%%%%%%%%%%%%%%%%%%%%%%%%%%%%%%%
% \section*{Acknowledgment}

% The preferred spelling of the word ``acknowledgment'' in American English is 
% without an ``e'' after the ``g.'' Use the singular heading even if you have 
% many acknowledgments. Avoid expressions such as ``One of us (S.B.A.) would 
% like to thank $\ldots$ .'' Instead, write ``F. A. Author thanks $\ldots$ .'' 
% \textbf{Sponsor and financial support acknowledgments are placed in the 
% unnumbered footnote on the first page, not here.}

% \balance
%%%%%%%%%%%%%%%%%%%%%%%%%%%%%%%%%%%%%%%%%%%%%%%%%%%%%%%%%%%%%%%%%%%%%%%%%%%%%%%%
\bibliographystyle{IEEEtran}
\bibliography{IEEEbibs} 
\vspace{-0.1cm}
%%%%%%%%%%%%%%%%%%%%%%%%%%%%%%%%%%%%%%%%%%%%%%%%%%%%%%%%%%%%%%%%%%%%%%%%%%%%%%%%

\vskip -2.5\baselineskip plus -1fil
%%%%%%%%%%%%%%%%%%%%%%%%%%%%%%%%%%%%%%%%%%%
\begin{IEEEbiography}[{\includegraphics[width=1in,height=1.25in,clip,keepaspectratio]{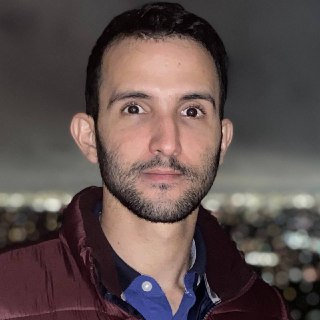}}]{Rodolfo Valiente}
is a Ph.D. candidate in Computer Engineering at the University of Central Florida (UCF). His research interests include connected autonomous vehicles (CAVs), reinforcement learning, computer vision, and deep learning with a focus on the autonomous driving problem. He received a M.Sc. degree from the University of Sao Paulo (USP) and his B.Sc. degree from the Technological University Jose Antonio Echeverria.
\end{IEEEbiography}
%%%%%%%%%%%%%%%%%%%%%%%%%%%%%%%%%%%%%%%%%%%
\vskip -2.5\baselineskip plus -1fil
%%%%%%%%%%%%%%%%%%%%%%%%%%%%%%%%%%%%%%%%%%%
\begin{IEEEbiography}[{\includegraphics[width=1in,height=1.25in,clip,keepaspectratio]{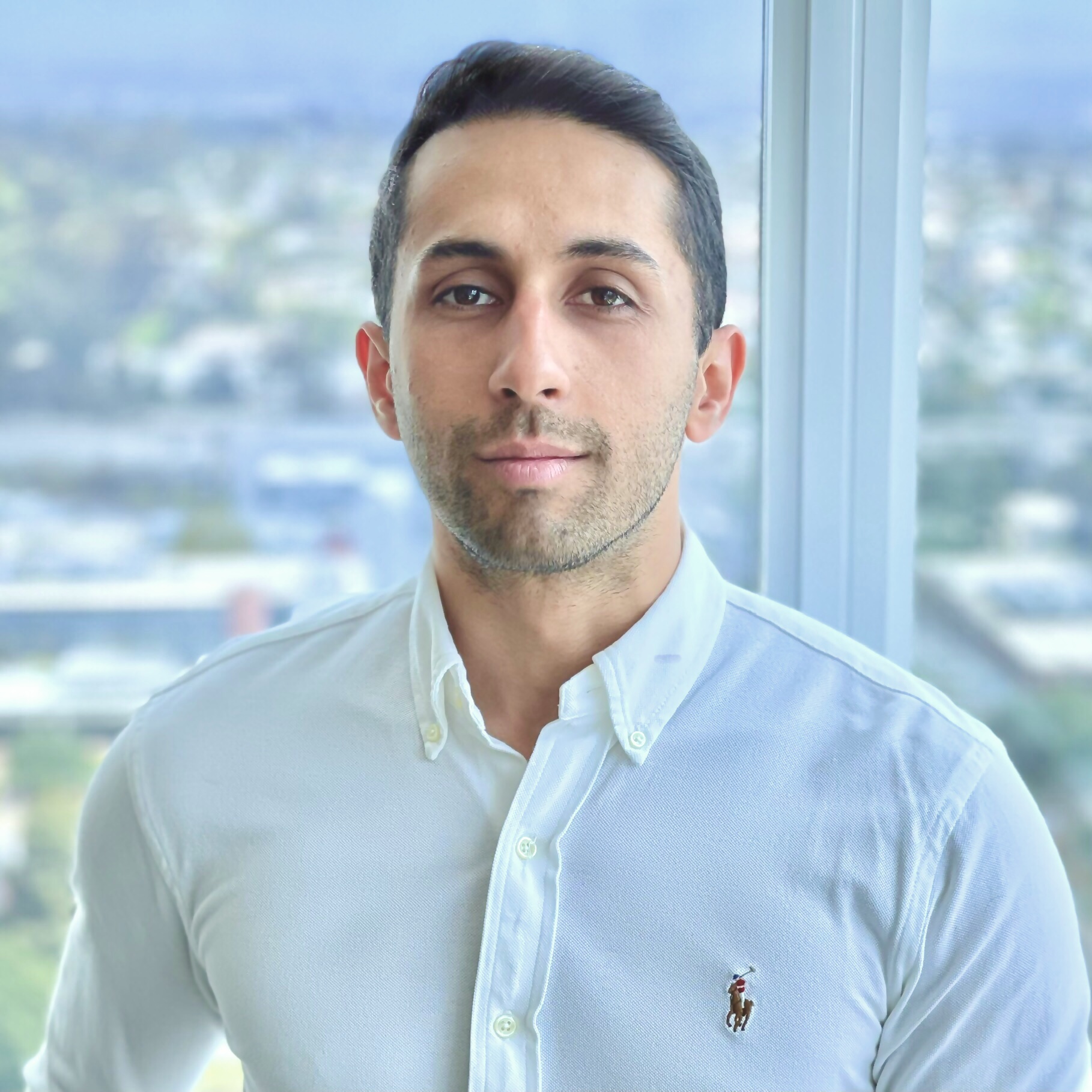}}]{Behrad Toghi}
is a Ph.D. candidate at the University of Central Florida. He received the B.Sc. degree in electrical engineering from Sharif University of Technology in 2016 and has worked as a research intern at Honda Research Institute, Mercedes-Benz R\&D North America and Ford Motor Company R\&D between 2018 and 2021. His work is mainly in the domains of prediction \& behavior planning for autonomous driving.
\end{IEEEbiography}
%%%%%%%%%%%%%%%%%%%%%%%%%%%%%%%%%%%%%%%%%%%
\vskip -2.5\baselineskip plus -1fil
%%%%%%%%%%%%%%%%%%%%%%%%%%%%%%%%%%%%%%%%%%%
\begin{IEEEbiography}[{\includegraphics[width=1in,height=1.25in,clip,keepaspectratio]{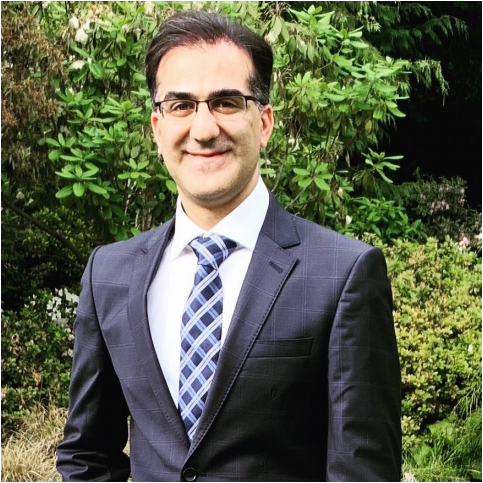}}]{Ramtin Pedarsani}
is an Assistant Professor in the ECE Department at the University of California, Santa Barbara. He received the B.Sc. degree in electrical engineering from the University of Tehran in 2009, the M.Sc. degree in communication systems from the Swiss Federal Institute of Technology (EPFL) in 2011, and his Ph.D. from the University of California, Berkeley, in 2015. His research interests include networks, game theory, machine learning, and transportation systems.
\end{IEEEbiography}
%%%%%%%%%%%%%%%%%%%%%%%%%%%%%%%%%%%%%%%%%%%
\vskip -2.5\baselineskip plus -1fil
%%%%%%%%%%%%%%%%%%%%%%%%%%%%%%%%%%%%%%%%%%%
\begin{IEEEbiography}[{\includegraphics[width=1in,height=1.25in,clip,keepaspectratio]{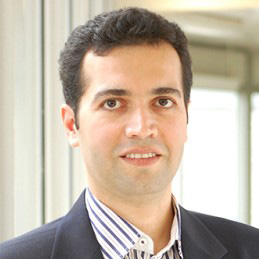}}]{Yaser P. Fallah} is an Associate Professor in the ECE Department at the University of Central Florida. He received the Ph.D. degree from the University of British Columbia, Vancouver, BC, Canada, in 2007. From 2008 to 2011, he was a Research Scientist with the Institute of Transportation Studies,
University of California Berkeley, Berkeley, CA, USA. His research, sponsored by industry, USDoT, and NSF, is focused on intelligent transportation systems and automated and networked vehicle safety systems.
\end{IEEEbiography}
%%%%%%%%%%%%%%%%%%%%%%%%%%%%%%%%%%%%%%%%%%%
% \vskip -2.5\baselineskip plus -1fil

\end{document}